\documentclass{article}

\usepackage{microtype}
\usepackage{graphicx}
\usepackage{subcaption}
\usepackage{booktabs} 

\usepackage{hyperref}

\usepackage[ruled,vlined]{algorithm2e}

\usepackage[preprint]{icml2026}

\usepackage{amsmath}
\usepackage{amssymb}
\usepackage{mathtools}
\usepackage{amsthm}
\usepackage{array}
\usepackage{bbding}
\usepackage{tcolorbox}
\usepackage{newfloat}
\usepackage{listings}
\usepackage{tabularx}
\usepackage{array} 
\usepackage{multirow}
\usepackage{booktabs}
\usepackage[capitalize,noabbrev]{cleveref}

\theoremstyle{plain}
\newtheorem{theorem}{Theorem}[section]

\newtheorem{lemma}[theorem]{Lemma}

\theoremstyle{definition}

\theoremstyle{remark}

\usepackage[textsize=tiny]{todonotes}

\icmltitlerunning{REST: Diffusion-based Real-time End-to-end Streaming Talking Head Generation via ID-Context Caching and Asynchronous Streaming Distillation}

\begin{document}

\twocolumn[
  \icmltitle{REST: Diffusion-based Real-time End-to-end Streaming Talking Head Generation via ID-Context Caching and Asynchronous Streaming Distillation}



  \icmlsetsymbol{equal}{*}

  \begin{icmlauthorlist}
    \icmlauthor{Haotian Wang}{equal,yyy}
    \icmlauthor{Yuzhe Weng}{equal,yyy}
    \icmlauthor{Jun Du}{yyy}
    \icmlauthor{Haoran Xu}{comp}
    \icmlauthor{Xiaoyan Wu}{comp}
    \icmlauthor{Shan He}{comp}
    \icmlauthor{Bing Yin}{comp}
    \icmlauthor{Cong Liu}{comp}
    \icmlauthor{Qingfeng Liu}{yyy,comp}
  \end{icmlauthorlist}

  \icmlaffiliation{yyy}{University of Science and Technology of China, China}
  \icmlaffiliation{comp}{iFLYTEK, China}

  \icmlcorrespondingauthor{Jun Du}{jundu@ustc.edu.cn}

  \icmlkeywords{Machine Learning, ICML}

  \vskip 0.3in
]



\printAffiliationsAndNotice{}  

\begin{abstract}
  Diffusion models have significantly advanced the field of talking head generation (THG). However, slow inference speeds and prevalent non-autoregressive paradigms severely constrain the application of diffusion-based THG models. In this study, we propose REST, a pioneering diffusion-based, real-time, end-to-end streaming audio-driven talking head generation framework. To support real-time end-to-end generation, a compact video latent space is first learned through a spatiotemporal variational autoencoder with a high compression ratio. Additionally, to enable semi-autoregressive streaming within the compact video latent space, we introduce an ID-Context Cache mechanism, which integrates ID-Sink and Context-Cache principles into key-value caching for maintaining identity consistency and temporal coherence during long-term streaming generation. Furthermore, an Asynchronous Streaming Distillation (ASD) strategy is proposed to mitigate error accumulation and enhance temporal consistency in streaming generation, leveraging a non-streaming teacher with an asynchronous noise schedule to supervise the streaming student. REST bridges the gap between autoregressive and diffusion-based approaches, achieving a breakthrough in efficiency for applications requiring real-time THG. Experimental results demonstrate that REST outperforms state-of-the-art methods in both generation speed and overall performance.
\end{abstract}

\section{Introduction}

Audio-driven Talking Head Generation (THG) aims to generate realistic talking head videos conditioned on input speech and reference images~\cite{thdsurvey1},  demonstrating significant research and application value in domains such as remote education~\cite{thdsurvey3}, virtual reality~\cite{thdsurvey2} and film production~\cite{thdsurvey4}. The lip-sync accuracy and naturalness of the generated videos serve as fundamental benchmarks for model performance~\cite{thdsurvey4}. Furthermore, due to the growing demands of real-time applications like human-computer interaction, the inference speed of THG models is garnering increasing research attention~\cite{thdsurvey2}.

Recent advances in talking head generation have been significantly driven by the introduction of diffusion models~\cite{diffusionsurvey}. Diffusion-based THG methods~\cite{emotivetalk,fantasytalking} demonstrate substantial improvements in both lip-sync accuracy and visual naturalness. Despite the improvements, two major limitations remain in current diffusion-based THG systems. First, end-to-end diffusion models suffer from slow inference speeds, typically requiring tens to hundreds of seconds to generate a five-second video~\cite{echomimicv3}. Second, most existing diffusion-based models operate in a non-streaming manner, which further increases latency for end users~\cite{emotivetalk}. Conversely, autoregressive (AR) THG methods~\cite{artalk} typically generate intermediate facial motion representations from speech in a sequential manner, which are then rendered into video via a separate rendering module, achieving lower latency and supporting streaming inference beyond diffusion-based methods. However, the output video quality of two-stage AR methods is limited by the expressiveness of the generated motion representations, resulting in generally inferior performance compared to end-to-end diffusion-based approaches~\cite{read}.

To bridge the gap between diffusion-based and AR-based THG methods in addressing key challenges in the field of THG, in this research we introduce a novel diffusion-based \textbf{R}eal-time \textbf{E}nd-to-end \textbf{S}treaming \textbf{T}alking head generation (\textbf{REST}) framework. We first utilize a variational autoencoder (VAE) with high spatiotemporal compression to learn a compact latent space, reducing the computational burden of end-to-end diffusion. Inspired by AR methods, we integrate a novel ID-Context Cache mechanism into our diffusion transformer backbone. By combining ID-Sink and Context-Cache, ID-Context Cache enables streaming inference while enhancing identity and temporal consistency. Furthermore, to address the quality degradation challenge inherent in streaming generation, an Asynchronous Streaming Distillation (ASD) scheme is proposed, which leverages latent spatiotemporal context generated by a non-streaming teacher model to supervise the streaming student model. 

\begin{figure}[t]
    \centering
    \includegraphics[width=1.00\linewidth]{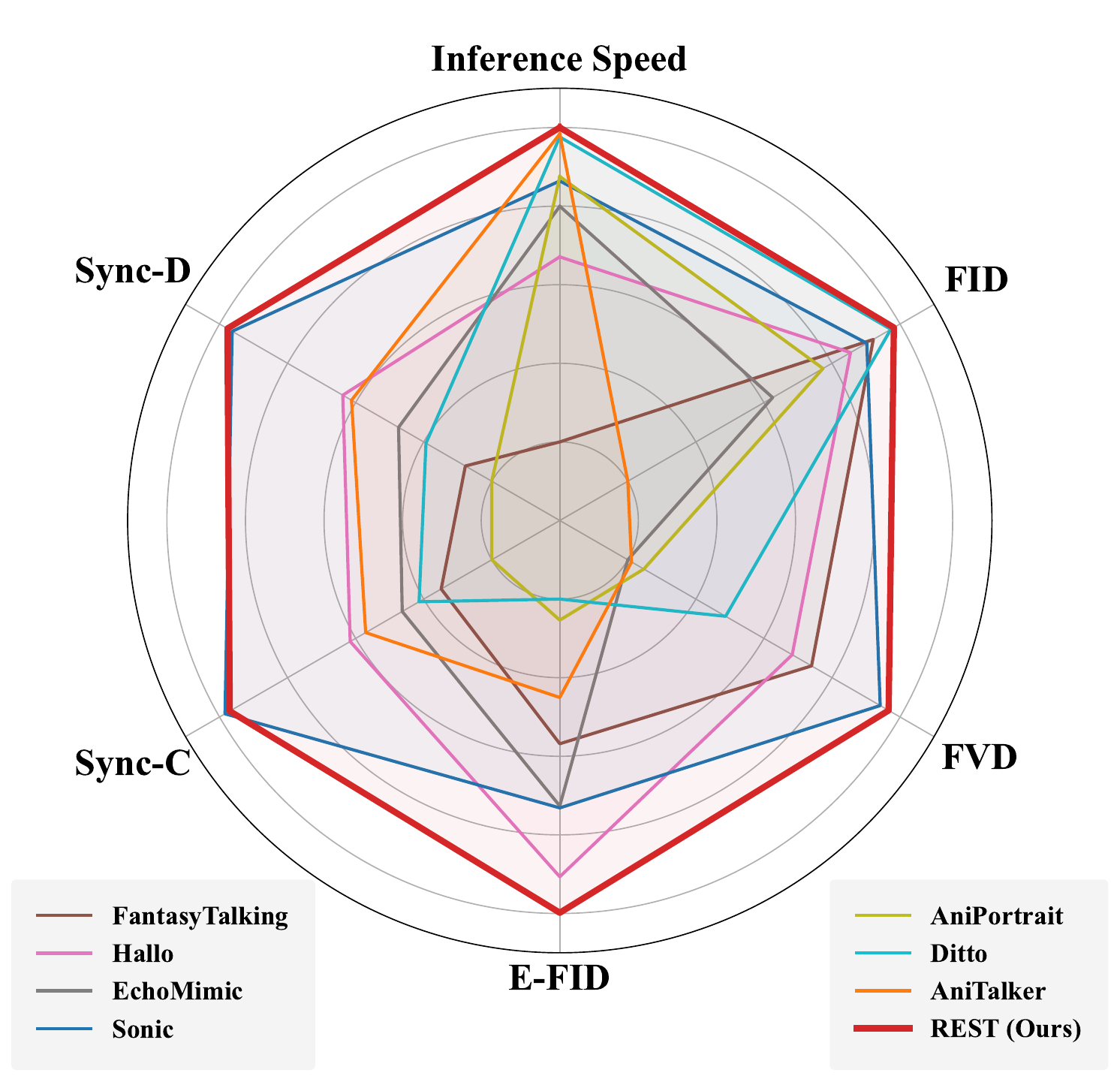}
    \caption{Performance benchmarking of the SOTA THG models.}
    \label{fig:radarcomparison}
\end{figure}

In summary, our contributions are as follows:

\begin{itemize}
    \item We propose REST, a novel framework for real-time, end-to-end, streaming talking head generation that effectively bridges the gap between diffusion and AR paradigms. To the best of our knowledge, REST is the first end-to-end method to achieve real-time streaming audio-driven talking head diffusion on a single GPU.
    
    \item We introduce a novel ID-Context Cache mechanism to facilitate high-quality streaming generation. By integrating ID-Sink and Context-Cache principles, our approach maintains high ID fidelity and temporal coherence across frames under streaming constraints.
    
    \item We further propose an Asynchronous Streaming Distillation (ASD) scheme to bridge the performance gap between streaming and non-streaming models by integrating information-theoretic alignment and kinematic constraints into a unified distillation objective.
    
\end{itemize}

\section{Related Work}

\subsection{Audio-driven Talking Head Generation}
Audio-driven Talking Head Generation (THG) is a cross-modal synthesis task aimed at generating a realistic talking head video conditioned on a reference image and a speech signal. Early efforts~\cite{wav2lip,diffusedheads} primarily focused on achieving lip-sync between the generated video and the input speech. Subsequent work~\cite{audio2head,sadtalker} expanded beyond lip-sync to improve the overall naturalness of the output, regarding natural facial expressions and motions. Recently, the success of diffusion models~\cite{ddpm,ddim} in image and video generation~\cite{diffusionsurvey} has revolutionized THG. End-to-end THG frameworks built upon pretrained diffusion models have demonstrated superior performance on lip-sync and naturalness. However, these improvements come at the cost of substantially increased latency~\cite{read}. Moreover, to ensure better temporal consistency, most current diffusion-based THG models typically adopt non-autoregressive paradigms~\cite{sonic}. As a result, users must wait for the entire video to be synthesized before viewing, which hinders real-world deployment. To address this limitation, we propose an end-to-end streaming THG framework based on ID-Context Cache and a streaming teacher-student learning scheme, achieving robust streaming generation with superior ID and temporal consistency.
\subsection{Fast Talking Head Generation}
Talking Head Generation differs from general video generation in its stronger demand for real-time performance, particularly in applications such as human-computer interaction and game production~\cite{thdsurvey2}. Conventional efficient THG often adopts a two-stage generation paradigm. In this approach, low-dimensional 2D or 3D motion representations~\cite{ditto, anitalker},  are utilized as video driving signals, and a renderer is employed to generate videos from these representations, accelerating inference by converting the high-dimensional latent generation task into a low-dimensional motion generation problem. Motion representation generation is primarily approached via diffusion or AR schemes. Motion-space diffusion methods such as AniTalker~\cite{anitalker} and Ditto~\cite{ditto} utilize lightweight diffusion models to predict motion representations, while AR-based approaches~\cite{artalk,multitalk} generate motions in a context-aware manner to enable streaming capabilities. However, two-stage methods are bottlenecked by the fidelity of generated motion representations, generally yielding naturalness inferior to end-to-end approaches~\cite{sonic,echomimicv3}. READ~\cite{read} represents the first attempt at real-time end-to-end talking head generation, achieving audio-visual alignment in a compressed latent space, but its non-autoregressive design still fails to support streaming real-time synthesis. In this work, we propose a pioneering diffusion-based end-to-end streaming THG model that successfully bridges AR and diffusion paradigms.

\begin{figure*}[t]
\centering
\includegraphics[width=1.00\textwidth]{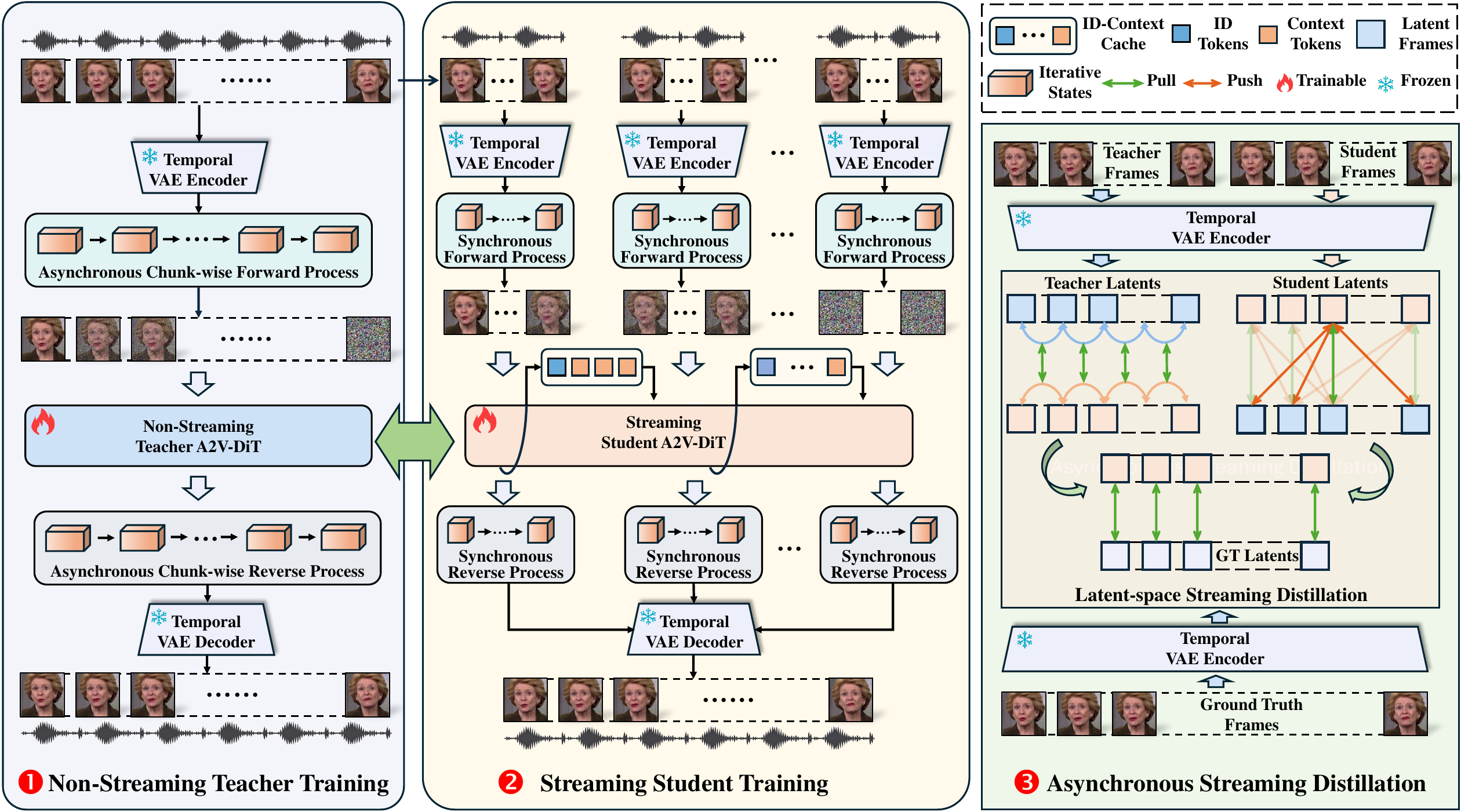} 
\caption{The overall framework of REST. During training, the non-streaming teacher model is first pre-trained with an asynchronous chunk-wise scheduler to provide a solid performance upper bound, shown in \textcircled{1}. Then the streaming student model with ID-Context Cache is trained under the guidance of the streaming teacher by ASD, as shown in \textcircled{2}. The principle of ASD is demonstrated in \textcircled{3}}.
\label{fig:overall}
\end{figure*}

\section{Methods}
The total framework of REST is shown in Fig.~\ref{fig:overall}. Sec.~\ref{sec:preliminaries} outlines the necessary preliminaries relevant to our work. Sec.~\ref{sec:idcontext} details the proposed model architecture of REST with ID-Context Cache. Sec.~\ref{sec:ASD} focuses on the model training methodology guided by the proposed ASD. And the final section gives an introduction to the AR-style inference methodology of the parametrized streaming REST model. 

\subsection{Preliminaries}
\label{sec:preliminaries}
\noindent\textbf{Task Definition.} 
Audio-driven THG task can be formulated as a mapping from a reference image $\mathit{\boldsymbol{I}_\text{ref}}\in\mathbb{R}^{H\times W\times3}$ and speech $\mathit{\boldsymbol{S}_{1:F_{a}}} $ to an output video $\mathit{\boldsymbol{X}}\in\mathbb{R}^{H\times W\times F\times D_\text{v}}$ ($D_\text{v}=3$), $F_{a}$ and $F$ denote the length of speech and video.

\noindent\textbf{Flow Matching.} 
The central idea of Flow Matching (FM)~\cite{flowmatching} is to learn a continuous-time vector field ${\mathit{\boldsymbol{v}}}({\mathit{\boldsymbol{Z}}}(t), t)$ that transports samples from a simple noise distribution $ {\mathit{\boldsymbol{Z}}}(t)$ to the target data distribution $ {\mathit{\boldsymbol{Z}}}(0)$~\cite{flowmatching1,flowmatching2}:
\begin{equation}
    {{\mathit{{d}}}{\mathit{\boldsymbol{Z}}}(t)} = {\mathit{\boldsymbol{v}}}({\mathit{\boldsymbol{Z}}}(t), t){dt}
\end{equation}
The forward process of FM defines a probability path from the original distribution $ {\mathit{\boldsymbol{Z}}}(0)$ to $ {\mathit{\boldsymbol{Z}}}(t)$. The process can be formulated when using Gaussian probability paths to add synchronous Gaussian noise at timestep $t$ to $ {\mathit{\boldsymbol{Z}}}(0)$:
\begin{equation}
    {\mathit{\boldsymbol{Z}}}(t) = (1-t){\mathit{\boldsymbol{Z}}}(0) + t\boldsymbol{\epsilon},~\boldsymbol{\epsilon}\sim\mathcal{N}(0,\boldsymbol{I})
\end{equation}
The training objective of FM is for the model $\theta$ to learn the correct vector field ${\mathit{\boldsymbol{u}}}({\mathit{\boldsymbol{Z}}}(t), t)$, as follows:
\begin{equation}
    \mathcal{L}_{\text{FM}}(\theta) = \mathbb{E}_{t,Z(t)\sim p_{t}}||{\mathit{\boldsymbol{v}}}({\mathit{\boldsymbol{Z}}}(t), t) - {\mathit{\boldsymbol{u}}}({\mathit{\boldsymbol{Z}}}(t), t)||^2
\end{equation}
where ${\mathit{\boldsymbol{u}}}({\mathit{\boldsymbol{Z}}}(t), t)$ is the target ground-truth vector field. 

\subsection{Streaming Diffusion with ID-Context Cache}
\label{sec:idcontext}
To achieve real-time streaming generation, we propose a framework comprising a temporal VAE and a DiT backbone integrated with the proposed ID-Context Cache mechanism.

\noindent \textbf{Temporal VAE for compact latent space.} For real-time efficiency in end-to-end diffusion, a compact latent space of videos is learned via a temporal VAE, which achieves a high compression ratio of 32×32×8 pixels per token, as inspired by LTX-Video~\cite{ltxvideo}. The principle can be formulated as follows:
\begin{equation}
    {\mathit{\boldsymbol{Z}}}(0) = \mathcal{E}_{\text{V}}({\mathit{\boldsymbol{X}}}(0)),~{\hat{\mathit{\boldsymbol{X}}}}(0) = \mathcal{D}_{\text{V}}({\mathit{\boldsymbol{Z}}}(0))
    \label{eq:vae}
\end{equation}
where ${\mathit{\boldsymbol{X}}}(0)\in\mathbb{R}^{H\times W\times F \times D_\text{v}}$ denotes the video sequence, and ${\mathit{\boldsymbol{Z}}}(0)\in\mathbb{R}^{h\times w\times f \times d_\text{v}}$ are the compressed video latents. $\mathcal{E}_{\text{V}}$ and $\mathcal{D}_{\text{V}}$ denote the encoder and decoder of VAE. 

\noindent \textbf{SpeechAE for compact speech latent space.} 
To facilitate cross-modal alignment within the compact latent space, we employ the SpeechAE architecture~\cite{read} to perform synchronous temporal compression on the input speech embedding $\mathit{\boldsymbol{S}_{1:F}}$. SpeechAE first integrates a Whisper-tiny encoder~\cite{whisper} for speech embedding extraction, as described below:
\begin{equation}
   \mathit{\boldsymbol{S}_{1:F}} = \mathcal{E}_{\text{Whisper}} (\mathit{\boldsymbol{A}_{1:F_{a}}} )
\end{equation}
Then the corresponding compact speech latent embedding $\mathit{\boldsymbol{E}\in\mathbb{R}^{f \times h_w\times d_{\text{A}}}}$ is generated by the SpeechAE encoder from $\mathit{\boldsymbol{S}}$, and reconstructed to $\hat{\mathit{\boldsymbol{S}}}$ through the decoder:
\begin{equation}
    {\mathit{\boldsymbol{E}}} = \mathcal{E}_{\text{A}}({\mathit{\boldsymbol{S}}}),~\hat{{\mathit{\boldsymbol{S}}}} = \mathcal{D}_{\text{A}}({\mathit{\boldsymbol{E}}})
\label{eqa:speechae}
\end{equation}
where $H_w$ and $h_w$ indicate the window sizes, $D_{\text{A}}$ and $d_{\text{A}}$ denote the hidden dimensions. Detailed architecture information is provided in Sec.~\ref{sec:backbone_details} of the Appendix.

\noindent \textbf{ID-Context Cache for semi-autoregressive diffusion.} 
To enable semi-autoregressive, context-aware generation in audio-to-video diffusion, we introduce a novel DiT-based backbone that implements the proposed ID-Context Cache mechanism. Our Streaming A2V-DiT framework comprises 28 transformer blocks, each integrating three core modules: Self-Attention with ID-Context Cache, 3D Full-Attention for text conditioning, and Frame-level 2D Cross-Attention for audio conditioning (please refer to Appendix Sec.~\ref{sec:backbone_details} for details). Theoretically, the ID-Context Cache reconfigures the attention topology to enable chunk-by-chunk semi-autoregressive inference, simultaneously enforcing ID fidelity and temporal consistency. This is realized through the two following theoretically motivated components. 

\begin{tcolorbox}[width=0.5\textwidth]
    \textbf{\textit{ID-Sink Principle}}: ID-Sink models the Key-Value (KV) embeddings of the reference image as a persistent sink across all chunks, establishing a global semantic anchor to preserve identity fidelity.

    \textbf{\textit{Context-Cache Principle}}: Context Cache approximates a continuous temporal flow by extending the effective temporal receptive field. It reconstructs local temporal dependencies by concatenating the KV embeddings of the preceding chunk with the current KV embedding during self-attention, thereby mitigating boundary discontinuities of streaming generation.
\end{tcolorbox}

The mathematical formulation of the ID-Context Cache is defined as follows. For the current chunk, the Query, Key, and Value at each block are computed as formulated below:
\begin{align}
\label{eq:selfattn1}
    & \mathit{\boldsymbol{Q}_{ij}}=\mathit{\boldsymbol{W}_j^Q} \cdot \mathit{\boldsymbol{H}_{ij}} \\
    & \mathit{\boldsymbol{K}_{ij}^{c}}=\mathit{\boldsymbol{W}_j^K} \cdot \mathit{\boldsymbol{H}_{ij}} \\
    & \mathit{\boldsymbol{V}_{ij}^{c}}=\mathit{\boldsymbol{W}_j^V} \cdot \mathit{\boldsymbol{H}_{ij}}
\end{align}
where $i$ denotes the $i$-th chunk and $j$ denotes the $j$-th block of A2V-DiT. The causal chain is then reconstructed by concatenating the static ID anchor with the dynamically updated contextual cache from the preceding chunk, following the proposed ID-Sink and Context-Cache principles:
\begin{align}
    & \mathit{\boldsymbol{K}_{ij}}= [\mathit{\boldsymbol{K}}_{0j}^{c} \mathbin{\|} \mathit{\boldsymbol{K}}_{(i-1)j}^{c} \mathbin{\|} \mathit{\boldsymbol{K}}_{ij}^{c}] \\
    & \mathit{\boldsymbol{V}_{ij}}= [\mathit{\boldsymbol{V}}_{0j}^{c} \mathbin{\|} \mathit{\boldsymbol{V}}_{(i-1)j}^{c} \mathbin{\|} \mathit{\boldsymbol{V}}_{ij}^{c}]
    \label{eq:selfattn2}
\end{align}
The self-attention is then operated on hidden states $\mathit{\boldsymbol{H}}_{ij}$ based on the refined causal dependency chain, as follows:
\begin{align}
    \mathit{\boldsymbol{H}}_{i(j+1)} = \mathit{\boldsymbol{H}}_{ij} + \text{Softmax}\left(\frac{\mathit{\boldsymbol{Q}}_{ij} \cdot \mathit{\boldsymbol{K}}_{ij}^\text{T}}{\sqrt{d_k}}\right) \cdot \mathit{\boldsymbol{V}}_{ij}
\label{eq:selfattn3}
\end{align}

Regarding cross-modal audio conditioning, the speech latent code $\boldsymbol{E}=[\boldsymbol{E}_1 \mathbin{\|} \cdots\mathbin{\|}\boldsymbol{E}_k]$ ($k$ denotes the number of chunks), which is temporally aligned with the video chunks, serves as the driving condition. By enforcing frame-level spatial cross-attention with a temporal receptive field of chunk length $l$, we achieve precise streaming audio-visual alignment within the compact latent space, formulated as:
\begin{equation}
  \mathit{\boldsymbol{H}}_{ij}^{\text{A}} = \mathit{\boldsymbol{H}}_{ij} +
  \text{CrossAttn}\left(\mathit{\boldsymbol{H}}_{ij},{\mathit{\boldsymbol{E}}_{i}}\right)
  \label{eq:speechattn}
\end{equation}
where $\mathit{\boldsymbol{H}}_{ij}, \mathit{\boldsymbol{H}}_{ij}^{\text{A}}\in \mathbb{R}^{h\times w\times f \times d}$ denotes the hidden states before and after frame-level spatial cross-attention of $j\!-\!\text{th}$ attention block. Our proposed design enables the semi-AR streaming generation of video latent chunks that are synchronized with the corresponding speech conditions. Additional analysis of the ID-Context Cache is provided in Sec.~\ref{sec:idcontext_details}.

\subsection{Training with Asynchronous Streaming Distillation}
\label{sec:ASD}
To mitigate autoregressive error propagation and enforce generation quality in chunk-by-chunk streaming diffusion, we introduce the Asynchronous Streaming Distillation (ASD) scheme. Theoretically, ASD aims to distill global temporal priors from a non-streaming teacher into a streaming student through asynchronous noise conditioning. 

\noindent\textbf{Asynchronous Non-streaming Teacher Training.} To emulate streaming dynamics within a non-streaming framework for establishing a theoretical upper bound of temporal consistency, we employ a chunk-wise asynchronous noise scheduler to guide the training of the non-streaming teacher. The teacher model receives the full latent sequence as input. We first concatenate the reference frame to the front of the initial video latents to provide speaker ID, as follows:
\begin{equation}
    \mathit{\boldsymbol{z}}_\text{R}=\mathcal{E}_{\text{V}}(\mathit{\boldsymbol{I}_\text{ref}}), {\mathit{\boldsymbol{Z}}}(0)=[\mathit{\boldsymbol{z}}_\text{R} \mathbin{\|} \mathit{\boldsymbol{Z}}_1(0)\mathbin{\|}\cdots\mathbin{\|}\mathit{\boldsymbol{Z}}_k(0)]
\end{equation}
Then the asynchronous add-noise process operates by applying a chunk-wise noise instance to each chunk, as follows: 
\begin{align}
\label{eq:addnoise}
    &\boldsymbol{t}\!=\![0 \mathbin{\|}\underbrace{t_1,\cdots,t_1}_{\text{1-th Chunk}} \mathbin{\|}\underbrace{t_2,\cdots,t_2}_{\text{2-th Chunk}}\mathbin{\|}\cdots\mathbin{\|}\underbrace{t_k,\cdots,t_k}_{\text{k-th Chunk}}] \\
    &\!{\mathit{\boldsymbol{Z}^\mathcal{T}}}(\boldsymbol{t})\!=\!(\boldsymbol{1}-\boldsymbol{t}) \odot {\mathit{\boldsymbol{Z}}}(0) + \boldsymbol{t} \odot \boldsymbol{\epsilon}
\end{align}
\noindent The final training objective of the teacher network parameters $\mathcal{T_{\theta}}$ is as follows, conditioned on the audio latents $\boldsymbol{E}$:
\begin{align}
    \boldsymbol{v} &= \boldsymbol{\epsilon} - {\mathit{\boldsymbol{Z}}}(0)\\
 \mathcal{L}_{\mathcal{T}_{\theta}} &= \mathbb{E}_{\boldsymbol{t},\boldsymbol{Z}^{\mathcal{T}}(\boldsymbol{t})}||\boldsymbol{v} - \mathcal{T}_{\theta}(\boldsymbol{Z}^{\mathcal{T}}(\boldsymbol{t}), \boldsymbol{E}, \mathit{\boldsymbol{z}_\text{R}},\boldsymbol{t})||^2
  \label{eq:teacher loss}
\end{align}
where $\boldsymbol{v}^\mathcal{T}$ denotes the correct velocity flow under the Gaussian probability path of the asynchronous add-noise process.

\noindent\textbf{Streaming Student Training.} 
Subsequently, to emulate inference-time causal dynamics, the student model is trained via a sequential chunk-wise paradigm. We apply the same add-noise paradigm as teacher model to obtain noisy latents $\boldsymbol{Z}^{\mathcal{S}}(\boldsymbol{t})$ from the initial latents $\boldsymbol{Z}(0)$, which are then partitioned into chunks $\boldsymbol{Z}^{\mathcal{S}}(\boldsymbol{t})=[\mathit{\boldsymbol{z}}_\text{R} \mathbin{\|} \mathit{\boldsymbol{Z}}_1(t)\mathbin{\|}\cdots\mathbin{\|}\mathit{\boldsymbol{Z}}_k(t)]$.
These chunks are processed sequentially by the student model with the proposed ID-Context Cache mechanism to execute semi-autoregressive generation. Also, the final training objective of the student network parameters $\mathcal{S_{\theta}}$ is as follows, conditioned on the audio latent chunks $\boldsymbol{E}=[\boldsymbol{E}_1 \mathbin{\|} \cdots\mathbin{\|}\boldsymbol{E}_k]$:
\begin{align}
    \boldsymbol{v} &= \boldsymbol{\epsilon} - \boldsymbol{Z}(0)\\
 \mathcal{L}_{\mathcal{S}_{\theta}} &= \mathbb{E}_{\boldsymbol{t},\boldsymbol{Z}^{\mathcal{S}}(\boldsymbol{t})}||\boldsymbol{v} - \mathcal{S}_{\theta}(\boldsymbol{Z}^{\mathcal{S}}(\boldsymbol{t}), \boldsymbol{E}, \mathit{\boldsymbol{z}_\text{R}},\boldsymbol{t})||^2
  \label{eq:student loss}
\end{align}

\noindent\textbf{Teacher-Student Learning with ASD.} 
There exists a fundamental information asymmetry between the teacher and the student. The non-streaming teacher model leverages global attention to achieve superior consistency, while the streaming student is limited by local causal attention under streaming constraints. To address this fundamental limitation, we propose ASD between the non-streaming teacher and the streaming student. From an information-theoretic perspective, we aim to maximize the average mutual information (MI) between the generative distributions of student and teacher, thereby introducing global consistency constraints under streaming conditions. This is achieved by maximizing the MI variational lower bound via frame-wise contrastive learning, specifically by minimizing the InfoNCE loss~\cite{infonce} of the output velocity flows ${\mathit{\boldsymbol{v}^{\mathcal{T}}}}=\{{\mathit{\boldsymbol{v}}}^\mathcal{T}_i\}_{i=0}^f$ and ${\mathit{\boldsymbol{v}^{\mathcal{S}}}}=\{{\mathit{\boldsymbol{v}}}^\mathcal{S}_i\}_{i=0}^f$ (proof in Appendix Sec.~\ref{sec:proveasdcon}). The contrastive objective can be formulated as follows:
\begin{equation}
  \mathcal{L}_{\text{CON}} = - \frac{1}{f} \sum_{i=1}^{f} \log \left( \frac{\exp \left( \mathrm{sim}(\boldsymbol{v}^\mathcal{S}_i, \boldsymbol{v}^\mathcal{T}_i) / \tau \right)}{ \sum\limits_{j=1}^{f} \exp \left( \mathrm{sim}(\boldsymbol{v}^\mathcal{S}_i, \boldsymbol{v}^\mathcal{T}_j) / \tau \right) } \right)
\end{equation}

Furthermore, from a kinematic perspective, to enforce inter-chunk temporal consistency, we introduce a smoothness matching objective that minimizes the divergence in second-order optical flow between the outputs of teacher and student to serve as a  global consistency constraint, formulated as:

\begin{align}
    \Delta^2 \boldsymbol{v}_{i} &= \boldsymbol{v}_{i+1}-2\boldsymbol{v}_{i}+\boldsymbol{v}_{i-1} \\
    \mathcal{L}_{\text{SMO}} &= \frac{1}{f-1}\sum\nolimits_{i = 1}^{f-1}  (\Delta^2 \boldsymbol{v}^\mathcal{S}_i-\Delta^2\boldsymbol{v}^\mathcal{T}_i)^2
\end{align}

The final diffusion objective for the streaming student is formulated as a reconstruction loss regularized by teacher-student flow contrast and smoothness matching constraints:
\begin{equation}
\mathcal{L}_{\text{ASD}}=\mathcal{L}_{\mathcal{S}_{\theta}}+\alpha\mathcal{L}_{\text{CON}}+\beta\mathcal{L}_{\text{SMO}}
\end{equation}

The analysis of loss complementarity is detailed in Sec.~\ref{sec:suppanalysis}. ASD serves to transfer global contextual knowledge from a non-streaming teacher to the streaming student, improving the stability and consistency of streaming generation.

\subsection{AR-style Streaming Inference}

Leveraging the ID-Context Cache and ASD strategies, we formulate an AR-style inference pipeline. The pipeline employs a dual-loop inference scheme, where the inner loop performs intra-chunk diffusion and the outer loop executes inter-chunk recursion, progressively propagating historical information and ID anchors through ID-Context Cache. The resulting velocity flow is resolved to the latent space by an FM scheduler and reconstructed to pixel space via the temporal VAE decoder. During inference, we incorporate the Joint CFG~\cite{read} mechanism for classifier-free guidance. The complete pipeline is detailed in Algorithm 1.

\begin{algorithm}[t]
    \DontPrintSemicolon
    \SetAlgoLined
    
    \SetKwInOut{Input}{\textbf{Input}}
    \SetKwInOut{Output}{\textbf{Output}}
    
    \caption{AR-style Streaming Inference}
    \label{power}

    \Input{
        Time schedule $\{T_1,\cdots,T_n\}$ (where $T_n=0$);\\
        Reference image $\boldsymbol{I}_\text{ref}$, $\boldsymbol{z}_\text{R}=\mathcal{E}_{\text{V}}(\boldsymbol{I}_\text{ref})$;\\
        Speech latents $\boldsymbol{E} \in \mathbb{R}^{L \times h_w \times d_a}$, chunk length $l$;\\
        Noise $\boldsymbol{\epsilon} = \{ \boldsymbol{\epsilon}_1,\!\cdots,\!\boldsymbol{\epsilon}_N \}\!\in\!\mathcal{N}(\mathbf{0}, \mathbf{I})$, $\boldsymbol{Z}(T_1)=\boldsymbol{\epsilon}$;\\
        Zeros ID-Context $\boldsymbol{C}=\{\boldsymbol{C}_1,\cdots,\boldsymbol{C}_k\}$ ($\boldsymbol{C}_0=\emptyset$).
    }
    \Output{
        Generated latents $\boldsymbol{Z}(0)\! \in\! \mathbb{R}^{h \times w \times L \times d_\text{v}}$ ($L$ frames).
    }

    \BlankLine
    $\boldsymbol{Z}(T_1) \gets \{\boldsymbol{Z}_1(T_1),\cdots,\boldsymbol{Z}_k(T_1)\}$ \tcp*[r]{Segment}\;
    
    \vspace{-1.2em} 
    
    $\boldsymbol{Z}_j(T_1) \gets \{\boldsymbol{z}_\text{R}, \boldsymbol{z}_{1+(j-1)l}(T_1),\cdots,\boldsymbol{z}_{jl}(T_1)\}$\;
    
    \vspace{0.1em}
    
    $\boldsymbol{E} \gets \{\boldsymbol{E}_1,\cdots,\boldsymbol{E}_k\}$\;

    \For{$i \leftarrow 1$ \KwTo $k$}{
        \tcp*[r]{Iterate over chunks}
        
        \For{$j \leftarrow 1$ \KwTo $n$}{
            \tcp*[r]{Iterate over timesteps} 
            $\;\;\boldsymbol{t}_j \gets [0, T_{j}, \cdots, T_{j}]$\; \\
            $\boldsymbol{Z}_i(T_{j+1}), \boldsymbol{C}_{i} \xleftarrow[\text{FM}]{\text{CFG}} \mathcal{S}_{\theta}\big(\boldsymbol{Z}_i(T_{j}), \boldsymbol{E}_i, \boldsymbol{C}_{i-1}, \boldsymbol{t}_j \big)$ \;
        }
    }
    
    $\boldsymbol{Z}(0) \gets \{\boldsymbol{Z}_1(T_{n}),\cdots,\boldsymbol{Z}_k(T_{n})\}$ \tcp*[r]{Update}
    \KwRet $\boldsymbol{X}(0) : \boldsymbol{X}(0)=\mathcal{D}_{\text{V}}(\boldsymbol{Z}(0))$\;
\end{algorithm}

\section{Experiments and Results}
\label{sec:results}
\subsection{Experimental Setup}
\noindent\textbf{Implementation Details.}
Experiments encompassing both training and inference are conducted on HDTF~\cite{hdtf} and MEAD~\cite{mead} datasets. 95\% data of both datasets is randomly allocated for training and the remaining 5\% for testing, ensuring identity-disjoint of dataset splitting. We employ a two-stage training strategy. In the first stage, the non-streaming teacher model is pre-trained at a resolution of $512\times512$ pixels with a learning rate of $1\times10^{-5}$. In the second stage, the streaming student model with ID-Context Cache is trained at a resolution of $512\times512$ pixels and 97 frames, with a learning rate of $1\times10^{-5}$ and a batch size of 1. All reported results use 8-step sampling and Joint-CFG with $\alpha=6.0$ unless specified. Both training and evaluation are performed on NVIDIA A100 GPUs.

\noindent\textbf{Evaluation Metrics.} Generation performance is assessed using several metrics. For visual quality, we employ the Fréchet Inception Distance (FID)~\cite{fid} for image-level realism between synthesized videos and reference images and the Fréchet Video Distance (FVD)~\cite{fvd} for frame-level realism between synthesized and ground-truth videos; lower values indicate better performance for both metrics. Lip synchronization is measured with SyncNet~\cite{syncnet}, where a higher Synchronization Confidence (Sync-C) and a lower Synchronization Distance (Sync-D) indicate superior alignment with speech input. We further use the Expression-FID (E-FID) metric from EMO~\cite{emo} to measure the expression divergence between synthesized and ground-truth videos, with lower values indicating a more faithful reproduction of expressions. Finally, we evaluate the efficiency of the diffusion backbone of each model by measuring the average runtime of the backbone per video (Runtime).

\noindent\textbf{Baselines.}
We benchmark our method against several open-source SOTA methods, including end-to-end diffusion methods such as Sonic~\cite{sonic}, EchoMimic~\cite{echomimic}, Hallo~\cite{hallo}, FantasyTalking~\cite{fantasytalking} and AniPortrait~\cite{aniportrait}, as well as motion-space diffusion methods like AniTalker~\cite{anitalker} and Ditto~\cite{ditto}. All comparisons are conducted on the same device using identical test data with the same length of 4.84s (121 frames) to ensure fair evaluation.

\subsection{Overall Comparison}
 

As shown in Tab.~\ref{tab:overall}, motion-space diffusion approaches such as Ditto and AniTalker generally demonstrate lower latency compared to end-to-end baselines. In contrast, our proposed end-to-end solution not only achieves a lower latency superior to all existing methods, but also pioneers streaming inference capabilities, which remain unattainable by existing alternatives. Beyond this efficiency, our approach achieves highly competitive performance, achieving either the best or second-best results across all metrics on both datasets, representing a significant breakthrough in real-time streaming talking head video generation.

\begin{table*}[t]
\centering
\begin{tabularx}{\textwidth}{@{}c|c|c|*{3}{>{\centering\arraybackslash}X}*{3}{>{\centering\arraybackslash}X}@{}}
\toprule
\textbf{Dataset} & \textbf{Method} & \textbf{Streaming} & \textbf{Runtime(s)} & \textbf{FID ($\downarrow$)} & \textbf{FVD ($\downarrow$)} & \textbf{E-FID ($\downarrow$)} & \textbf{Sync-C ($\uparrow$)} & \textbf{Sync-D ($\downarrow$)} \\ 
\midrule
\multirow{9}{*}{HDTF} & Fantasy &\XSolidBrush & 896.089 & 16.489 & 315.291 & 1.232 & 5.138 & 10.349 \\
& Hallo &\XSolidBrush & 212.002 & 15.929 & 315.904 & {\underline{0.931}} & 6.995 & 7.819 \\
& EchoMimic &\XSolidBrush & 124.105 & 18.384 & 557.809 & \textbf{0.927} & 5.852 & 9.052 \\
& Sonic &\XSolidBrush & 83.584 & 16.894 & \underline{245.416} & 0.932 & \textbf{8.525} & \textbf{6.576} \\
& AniPortrait &\XSolidBrush & 76.778 & 17.603 & 503.622 & 2.323 & 3.555 & 10.830 \\
& Ditto &\XSolidBrush & 17.974 & {\underline{15.440}} & 399.965 & 2.659 & 5.458 & 9.565 \\
& AniTalker &\XSolidBrush & \underline{13.577} & 39.155 & 514.388 & 1.523 & 5.838 & 8.736 \\
& Ours &\CheckmarkBold & \textbf{4.416} & \textbf{14.597} & \textbf{219.870} & \underline{0.931} & \underline{8.335} & \underline{6.701} \\ 
\midrule
\multirow{9}{*}{MEAD} & Fantasy &\XSolidBrush & 896.089 & 46.617 & 257.077 & 1.510 & 4.536 & 10.699 \\
& Hallo &\XSolidBrush & 212.002 & 52.300 & 292.983 & {\underline{1.171}} & 6.014 & 8.822 \\
& EchoMimic &\XSolidBrush & 124.105 & 65.771 & 667.999 & 1.448 & 5.482 & 9.128 \\
& Sonic &\XSolidBrush & 83.854 & 47.070 & \textbf{218.308} & 1.434 & {\underline{7.501}} & \underline{7.831} \\
& AniPortrait &\XSolidBrush & 76.778 & 54.621 & 531.663 & 1.669 & 1.189 & 13.013 \\
& Ditto &\XSolidBrush & 17.974 & \textbf{45.403} & 349.860 & 1.941 & 5.199 & 9.595 \\
& AniTalker &\XSolidBrush & \underline{13.577} & 95.131 & 621.528 & 1.553 & 6.638 & 8.184 \\
& Ours &\CheckmarkBold & \textbf{4.416} & {\underline{46.540}} & {\underline{237.521}} & \textbf{1.064} & \textbf{7.632} & {\textbf{7.573}} \\ 
\bottomrule
\end{tabularx}
\caption{Overall comparisons on HDTF and MEAD. ``$\uparrow$'' indicates better performance with higher values, while ``$\downarrow$'' indicates better performance with lower values. The best results are \textbf{bold}, and the second-best results are \underline{underlined}.}
\label{tab:overall}
\end{table*}

\subsection{Ablation Study}
\label{sec:ablation}
To assess the contribution of each component in our proposed method, several ablation studies are carried out.

\begin{table}[t]
\centering
\resizebox{\linewidth}{!}{
\begin{tabular}{c|ccc}
\toprule
\textbf{Method} & \textbf{FID} ($\downarrow$) & \textbf{FVD} ($\downarrow$) & \textbf{Sync-C} ($\uparrow$) \\
\midrule
Full ID‑Context & \textbf{14.597} & \textbf{219.870} & \textbf{8.335}  \\
w/o ID-Sink & 19.362 & 294.692 & 8.345   \\
w/o Context-Cache & 19.656 & 271.508  & 6.909 \\
\bottomrule
\end{tabular}
}
\caption{Ablation results of ID-Context Cache on HDTF dataset.} 
\label{tab:idcontext}
\end{table}

\begin{figure}[t]
    \centering
    \includegraphics[width=1.00\linewidth]{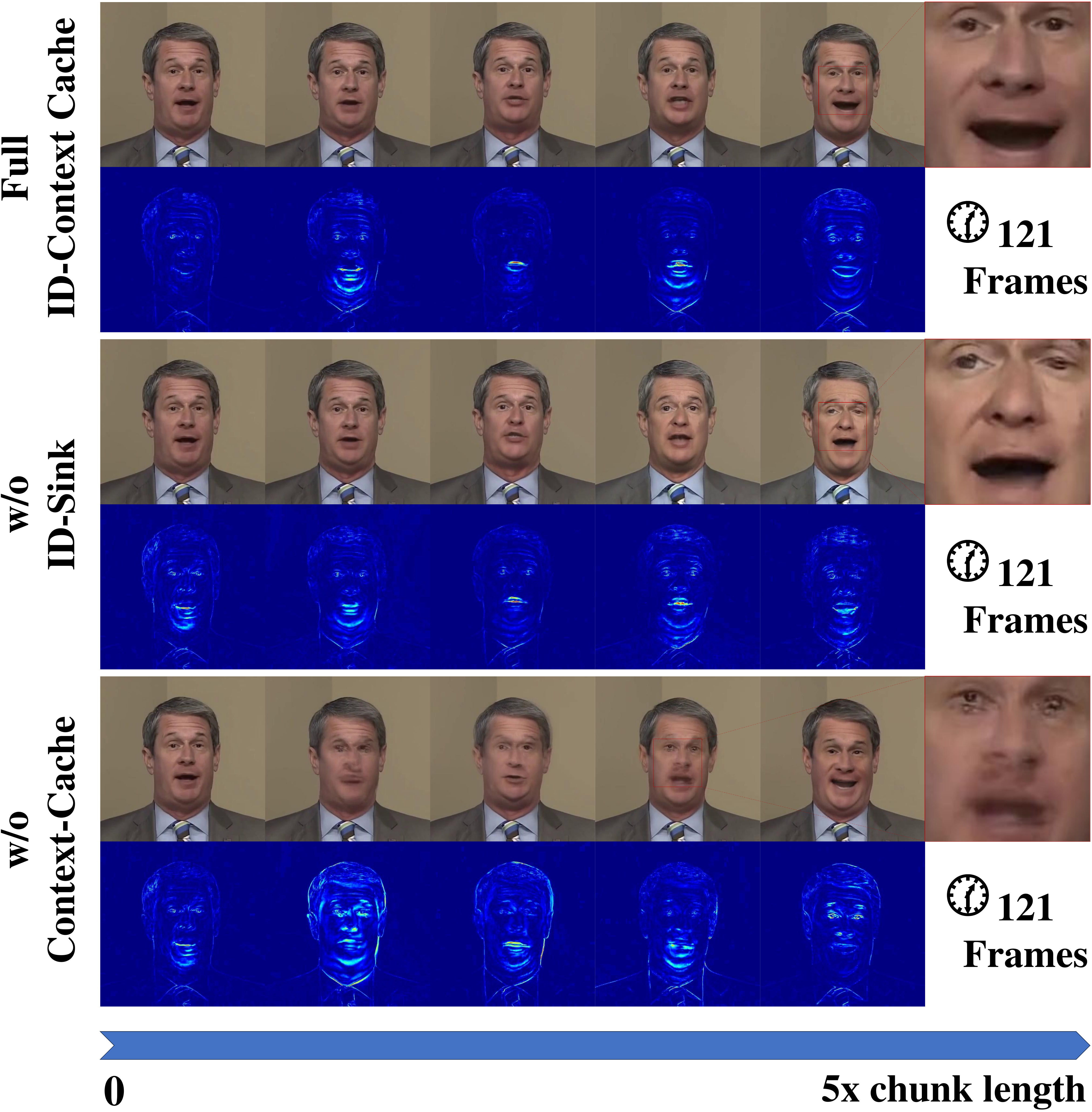}
    \caption{Visualization of the ablation results of ID-Context Cache.}
    \label{fig:ablation_idcontext}
\end{figure}

\noindent\textbf{Ablation Study on ID-Context Cache.} 
To investigate the specific contributions of the ID-Sink and Context-Cache principles within our proposed ID-Context Cache methodology, we conduct an ablation study under different experimental configurations, as follows:
\begin{itemize}
    \item \textbf{Full ID‑Context Cache}: Our complete strategy including both the ID‑Sink and Context‑Cache principles.
    \item \textbf{w/o ID-Sink}: Configuration excludes the ID‑Sink principle, retaining only the Context‑Cache principle.
    \item \textbf{w/o Context-Cache}: Configuration excludes the Context‑Cache, retaining only the ID-Sink principle.
\end{itemize}
The inference segment length is set to five times the training chunk length (121 frames). The performance of each configuration is evaluated on the HDTF dataset across FID, FVD, and SyncNet metrics. As detailed in Tab.~\ref{tab:idcontext}, ablating the ID-Sink mechanism has a minimal impact on Sync-C but significantly worsens FID and FVD, indicating that suboptimal ID consistency leads to semantic distortion in generated videos. In contrast, ablating the Context-Cache mechanism results in a clear deterioration in Sync-C, accompanied by declines in FID and FVD, which can be attributed to weakened temporal consistency. To further investigate these effects, we conducted a visual analysis by uniformly sampling frames at aligned timestamps and computing difference heatmaps between adjacent frames. The results are shown in Fig.~\ref{fig:ablation_idcontext}, where the inference length is 121 frames.

As observed, the generation results of w/o ID‑Sink configuration exhibit progressively deteriorating ID consistency over time, accompanied by a degradation in fine-grained details (e.g., the eyes in the zoomed-in views). Moreover, the overall color tone gradually degrades as inference proceeds, and the error heatmap shows high‑error regions not only on the subject but also across the background, indicating suboptimal ID consistency. Conversely, the w/o Context-Cache configuration suffers from compromised motion smoothness, characterized by abrupt motion discontinuities at chunk boundaries (see zoomed-in views). In addition, the difference heatmap displays overall higher error values distributed across the entire figure, confirming inferior temporal consistency and motion smoothness. In contrast, the Full ID-Context Cache, which integrates both principles, achieves superior performance in both identity preservation and motion smoothness. The zoom‑in details are clearly superior to those of the ablated variants. Notably, in the difference heatmaps of the results of Full ID‑Context Cache, high values are concentrated solely on dynamic facial regions (e.g., eyes and mouth), while the background and overall head structure exhibit minimal error. These visualizations confirm the crucial roles of ID-Sink and Context-Cache in enhancing identity consistency and temporal smoothness, demonstrating the effectiveness of our proposed method.

\noindent\textbf{Ablation Study on ASD.} 
To validate the effectiveness of our proposed ASD strategy and the contribution of each loss term, we conduct a quantitative ablation study using the following configurations:
\begin{itemize}
    \item \textbf{Full ASD}: Training with the complete ASD strategy.
    \item \textbf{w/o Smooth}: Training without Smooth loss.
    \item \textbf{w/o Contrastive}: Training without Contrastive loss.
    \item \textbf{w/o ASD}: Training without the ASD strategy.
\end{itemize}


\begin{table}[t]
\centering
\begin{tabular}{c|ccc}
\toprule
\textbf{Method} & \textbf{FID} ($\downarrow$) & \textbf{FVD} ($\downarrow$) & \textbf{Sync-C} ($\uparrow$) \\
\midrule
Full ASD & \textbf{14.597} & \textbf{219.870} & \textbf{8.335}  \\
w/o Smooth & 14.646 & 221.525 & 8.264   \\
w/o Contrastive & 15.278 & {219.881}  & 8.190 \\
w/o ASD & 15.950 & 228.815  & 7.970 \\
\bottomrule
\end{tabular}
\caption{Ablation results of ASD on HDTF dataset.} 
\label{tab:ASD}
\end{table}

We report the FID, FVD, and SyncNet scores of each configuration on the HDTF dataset, as shown in Tab~\ref{tab:ASD}. As indicated by the results, removing the Smooth loss while retaining the Contrastive loss for teacher-student distillation leads to a deterioration in the FVD metric, whereas FID and Sync-C scores show only marginal degradation. This indicates that the Smooth loss plays a crucial role in enhancing temporal coherence and preserving the overall semantic consistency of the generated videos. Conversely, removing the Contrastive loss and retaining only the Smooth loss results in noticeable deterioration in both Sync-C and FID metrics, with minimal impact on FVD. This demonstrates that the Contrastive loss primarily contributes to improving the overall visual quality and audio-visual alignment of the student model during streaming generation. Furthermore, complete removal of the ASD strategy shows a substantial performance drop across all three metrics compared to the Full ASD configuration. These results validate the significant efficacy of our non-streaming to streaming distillation strategy in improving both video quality and temporal consistency of semi-autoregressive streaming generation.

\subsection{Case Study}

\begin{figure}[t]
    \centering
    \includegraphics[width=1.00\linewidth]{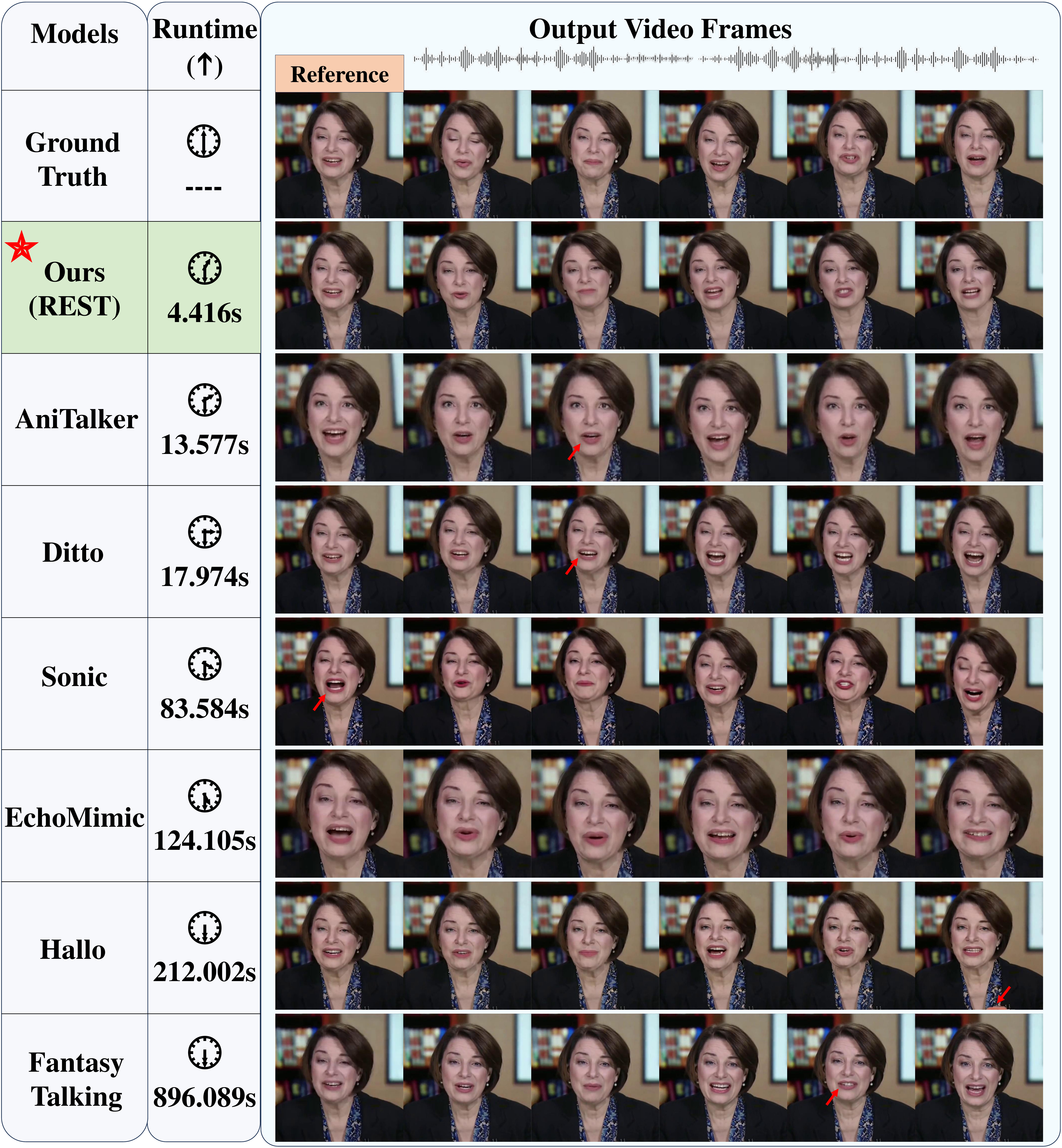}
    \caption{Case study of SOTA THG methods.}
    \label{fig:casestudy}
\end{figure}

To provide a more intuitive comparison of the generation quality among SOTA THG methods, we present a visual analysis of a representative test case. Conditioned on the identical reference image and driving audio, we utilize each model to generate talking head videos of equal length of 121 frames. Key frames at identical timestamps are sampled for visual comparison, as illustrated in Fig.~\ref{fig:casestudy}, with the inference runtime of the backbone network for each model provided.

Visual analysis of the representative case reveals distinct performance characteristics among THG methods. Motion-space diffusion methods such as AniTalker and Ditto achieve low runtime latency, yet exhibit poor lip-sync, with visible mismatches between generated and ground truth frames at corresponding timestamps (indicated by red arrows), underscoring the limitations of their two-stage training pipelines. End-to-end models, including Sonic and EchoMimic, show improved lip alignment, though occasional inaccuracies persist. Meanwhile, Hallo suffers from noticeable artifacts in later frames, attributable to its inferior generative quality. In contrast, our model operates with minimal inference time while maintaining accurate lip synchronization under real-time streaming constraints. The generated lip motions are closely aligned with the ground truth, and the outputs exhibit natural head movements and facial expressions. Together, these results demonstrate that our approach achieves an optimal balance between efficiency and perceptual quality, confirming the effectiveness of the proposed method.

\section{Conclusion}

This study addresses the critical and fundamental challenges of slow inference speed and the lack of streaming capability in existing diffusion-based talking head generation (THG) models by proposing REST, a novel diffusion-based, real-time end-to-end streaming framework for audio-driven talking head generation. A key innovation of REST is the introduction of the ID-Context Cache mechanism, which efficiently caches ID-related information and historical context associated with motion continuity to enable semi-autoregressive contextual streaming inference.  Furthermore, to tackle the quality and consistency bottlenecks inherent in streaming scenarios, we further propose the asynchronous streaming distillation (ASD) strategy. This approach first employs an asynchronous noise scheduler to simulate streaming conditions, aligning the non-streaming teacher model with the streaming student model. Through the combination of information-theoretic alignment and kinematic constraints, the global inter-frame dependency knowledge can be effectively distilled from the teacher model into the student model, thereby enhancing the long-term generation quality and temporal consistency of the student model under streaming constraints. In general, REST effectively bridges autoregressive and diffusion models, achieving SOTA performance in both speed and quality under real-time streaming conditions. Extensive ablation studies further validate the effectiveness of REST. 

\section*{Impact Statement}

This work aims to advance the field of Generative AI. Acknowledging the potential societal consequences of our work, we provide a detailed discussion of these impacts, alongside our corresponding mitigation strategies in Sec.~\ref{sec:ethicalconsideration}.


\bibliography{example_paper}
\bibliographystyle{icml2026}

\newpage
\appendix
\onecolumn
\section{Dataset Details.}
\label{sec:dataset}
\subsection{Benchmark Datasets}
To ensure the generalizability and expressive fidelity of our proposed framework, we conduct extensive experiments on two seminal benchmarks in talking head generation: the High-Definition Talking Face (HDTF)~\cite{hdtf} dataset and the Multi-view Emotional Audio-visual Dataset (MEAD)~\cite{mead}. The following subsection elucidates the statistical characteristics of these datasets and justifies their adoption as the primary testbeds for validating both the visual quality and emotional consistency of our model.

\noindent \textbf{High-Definition Talking Face (HDTF) Dataset.} The HDTF dataset~\cite{hdtf} serves as a prominent benchmark for high-resolution, in-the-wild talking head generation. Curated from online media, it comprises 15.8 hours of footage featuring 362 distinct subjects and over 10,000 unique utterances, ensuring a rich distribution of identities and phonetic content. We leverage HDTF in our framework primarily for its two distinguishing attributes: \begin{itemize} \item \textbf{Visual and Temporal Quality:} The dataset is predominantly composed of 720p and 1080p frontal-view videos. These high-definition sequences provide intricate facial textures and strictly synchronized lip motions, offering a robust supervision signal for learning identity preservation and precise lip-sync capabilities. \item \textbf{Domain Variability:} Unlike controlled lab environments, HDTF exhibits significant variance in illumination, background clutter, and demographics (spanning diverse ages and ethnicities). This real-world domain complexity is essential for evaluating our model's robustness and generalization to unconstrained scenarios. \end{itemize}

\noindent \textbf{Multi-view Emotional Audio-visual Dataset (MEAD).} Complementing the in-the-wild nature of HDTF, the MEAD dataset~\cite{mead} is a large-scale dataset specifically tailored for emotional talking head generation and multi-view consistency. It features high-quality recordings of 60 actors, offering a structured environment for analyzing affective facial dynamics. We incorporate MEAD into our experimental design to leverage its unique characteristics: \begin{itemize} \item \textbf{Fine-Grained Emotional Granularity:} MEAD provides annotations for 8 distinct emotion categories (e.g., happy, sad, angry) across 3 intensity levels. This hierarchical emotional labeling is critical for training and evaluating our model's capacity to synthesize expressive facial animations with nuanced affective dynamics. \item \textbf{Controlled Multi-View Environment:} In contrast to the variable backgrounds of HDTF, MEAD is captured in a strictly controlled studio setting with clean backgrounds across 7 distinct camera angles. This noise-free, high-quality data isolates facial motion from environmental interference, facilitating the learning of pure motion patterns and view-dependent generation. \end{itemize}

\subsection{Data Pre-processing} 
\label{sec:preprocess}
To ensure robust model training and rigorous evaluation, we implemented a systematic data processing pipeline applied uniformly to both the HDTF~\cite{hdtf} and MEAD~\cite{mead} datasets. This pipeline comprises three integral stages: visual data standardization, acoustic feature extraction, and sample quality filtering.

\noindent\textbf{Visual Data Standardization.} For the visual modality, our objective is to enforce spatiotemporal consistency across diverse data sources. We first standardize the temporal resolution of all video clips to 25 fps using FFmpeg. To achieve spatial alignment, we employ the OpenFace toolkit~\cite{openface} to detect 68 facial landmarks for each frame. To prevent jitter and ensure stable framing, rather than cropping frame-by-frame, we compute a static bounding box based on the global extrema of the landmarks across the entire video sequence. The videos are then cropped and resized to a unified $1:1$ aspect ratio. Finally, the processed frames are re-encoded into video sequences to optimize the I/O efficiency.

\noindent\textbf{Acoustic Feature Preparation.} For the acoustic modality, we first standardize raw audio streams into 16 kHz mono-channel waveforms. However, simply utilizing the final-layer output of pre-trained models often fails to capture the full spectrum of audio-visual correlations, including the phonetic nuances required for precise lip synchronization and the prosodic information essential for emotional expression. To address this limitation, we diverge from standard practices and extract hierarchical semantic speech representations~\cite{acm2023} from the Whisper-tiny encoder~\cite{whisper}, leveraging the complementarity of different feature levels from the self-supervised pretrained speech encoder. Simultaneously, to enhance the temporal coherence of the generated facial dynamics, we organize these hierarchical features into sliding temporal windows, which provides a broader receptive field for the spatial attention mechanism of the backbone model, enabling the model to account for past and future context. Formally, this extraction process can be defined as:
\begin{equation}
\mathbf{S}_{1:N} = \mathcal{E}_{\text{Whisper}} (\mathbf{A}_{1:N{a}})
\end{equation}
\noindent where $\mathbf{A}_{1:N_{a}}$ denotes the standardized input audio stream. The resulting acoustic speech representation is initially formulated as $\mathbf{S} \in \mathbb{R}^{N \times H_w \times H_l \times D_{\text{A}}^{\text{base}}}$. Here, $N$ corresponds to the video frame count (at 25 fps), and $H_w=10$ denotes the temporal window size. Crucially, $H_l=5$ represents the number of selected hierarchical layers, and $D_{\text{A}}^{\text{base}}=384$ is the fundamental hidden dimension of the Whisper encoder. In the final processing step, to effectively integrate these multi-level representations for model input, we concatenate the features along the channel dimension. Consequently, the effective feature dimension becomes $D_{\text{A}} = H_l \times D_{\text{A}}^{\text{base}} = 1920$. This unified representation $\mathbf{S}_{1:N} \in \mathbb{R}^{N \times H_w \times D_{\text{A}}}$ serves as the comprehensive driven condition for talking head video generation.

\noindent\textbf{Sample Quality Filtering.}To guarantee the fidelity of the training signal, we introduce a strict filtering protocol to discard samples containing occlusions or visual noise. This stage specifically addresses two degradation factors:
\begin{itemize}
\item \textbf{Occlusion Elimination:} Hand movements covering the face can introduce significant artifacts. We utilize MediaPipe~\cite{mediapipe} for hand keypoint detection, filtering out any video clips where the hand detection confidence exceeds a threshold of 0.8.
\item \textbf{Text Artifact Removal:} Visual overlays, such as subtitles, act as noise that impedes the learning of natural facial textures. We employ the pre-trained CRAFT~\cite{craft} text detector to scan the lower regions of video frames, automatically excluding any clips with detected text regions.
\end{itemize}

\section{Model Details}
\label{sec:architecture}
In this section, we provide a detailed description of the architecture of our proposed framework, including the self-attention mechanism with ID-Context Cache and the detailed design of the Streaming A2V-DiT backbone.

\subsection{Detailed Analysis of the ID-Context Cache Mechanism} 
\label{sec:idcontext_details}
\begin{figure}[t]
    \centering
    \includegraphics[width=1.0\linewidth]{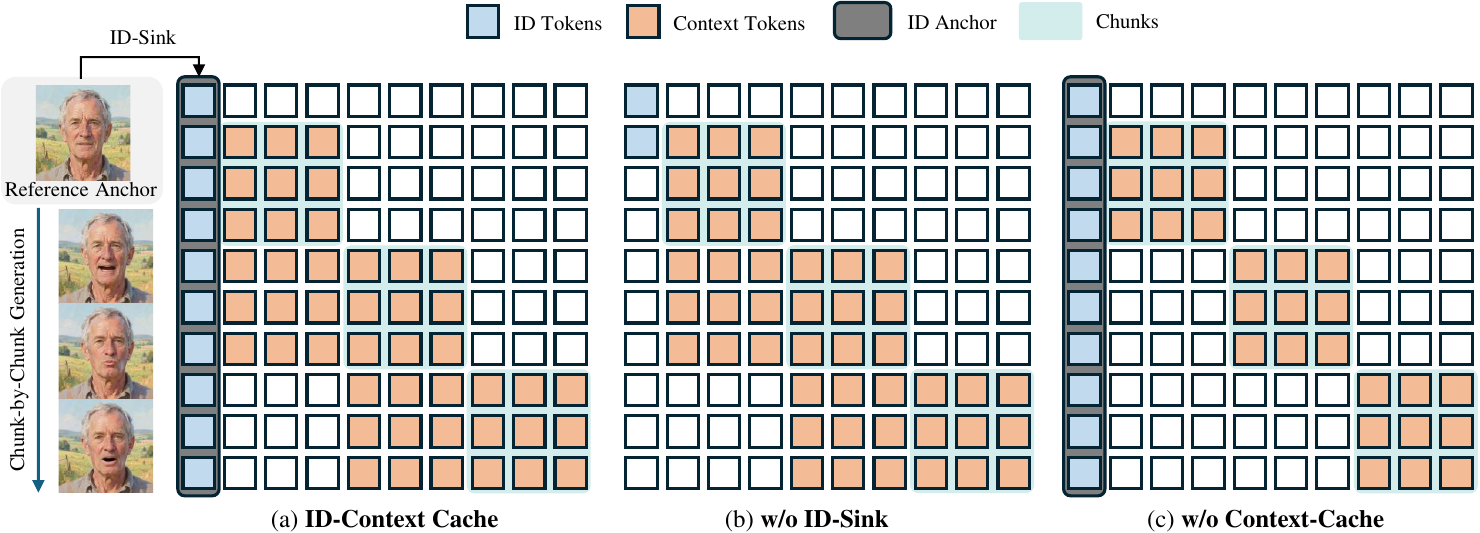}
    \caption{Comparison of self-attention mechanisms under different caching strategies. (a) ID-Context Cache: ID tokens derived from the reference image serve as persistent anchors across all chunks. Starting from the second chunk, the self-attention receptive field encompasses frames from both the current and the immediately preceding chunk. (b) w/o ID-Sink: The ID anchor is discarded after the first chunk. (c) w/o Context-Cache: For every chunk, the receptive field for self-attention is restricted to frames within the current chunk.}
    \label{fig:attention}
\end{figure}

This section serves as a supplement to Sec.~\ref{sec:idcontext}, elaborating on the underlying principles of the proposed ID-Context Cache self-attention mechanism and distinguishing the specific roles of its two constituent components: ID-Sink and Context-Cache. Fig.~\ref{fig:attention} presents a comparative schematic of the attention mechanisms under three configurations: the full ID-Context Cache (a), the removal of the ID-Sink (b), and the removal of the Context-Cache (c).

As illustrated in Fig.~\ref{fig:attention} (a), the self-attention mechanism with ID-Context Cache principle utilizes ID tokens extracted from the reference image as persistent anchors across all video chunks. This establishes a globally invariant semantic reference along the temporal axis, compelling the model to strictly adhere to the original speaker's identity throughout chunk-by-chunk generation. Starting from the second chunk, the Context-Cache principle effectively expands the self-attention receptive field to encompass both the current and the immediately preceding window, thereby integrating historical context into the latent generation of the current chunk. In our implementation, the chunk length is set to 4 latent frames (including one ID anchor frame), with an attention temporal window of 7 frames, covering all frames in both the current and the previous chunk. 

Conversely, the w/o ID-Sink variant disconnects from the reference identity anchor starting from the second chunk, relying solely on historical information for identity consistency. As corroborated by the experimental results in Tab.~\ref{tab:idcontext} and Fig.~\ref{fig:ablation_idcontext}, the absence of the ID-Sink leads to gradual identity drift and deviation from the semantic anchor as generation progresses. Similarly, the w/o Context-Cache variant fails to extend the temporal receptive field using history context, limiting the latent frames to attend only within the current chunk. Results in Tab.~\ref{tab:idcontext} and Fig.~\ref{fig:ablation_idcontext} confirm that w/o Context-Cache causes severe motion distortion artifacts at inter-chunk boundaries, reflecting poor temporal consistency.

\subsection{Implementation Details of the Streaming A2V-DiT}
\label{sec:backbone_details}

\begin{figure}[t]
    \centering
    \includegraphics[width=0.78\linewidth]{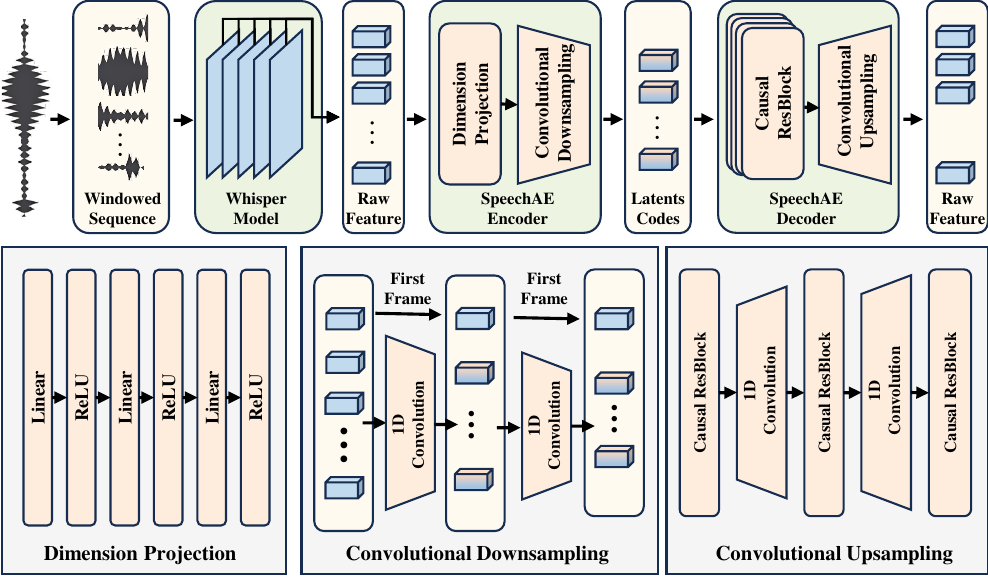}
    \caption{Detailed architecture of SpeechAE. SpeechAE employs an encoder-decoder architecture. Hierarchical raw features are encoded into speech latent codes by SpeechAE Encoder and are decoded back to raw features by SpeechAE Decoder.}
    \label{fig:speechae}
\end{figure}

This section supplements Sec.~\ref{sec:idcontext} of the main text by providing a comprehensive exposition of the diffusion backbone architecture. We detail the structural design of the Temporal VAE for spatio-temporal video latent compression, the SpeechAE module for speech latent encoding, and the Cross-Modal Diffusion Transformer (A2V-DiT).

\noindent\textbf{Detailed Temporal VAE Architecture.}
In this study, our Temporal VAE is initialized from the Video-VAE of LTX-Video~\cite{ltxvideo}. It applies a spatio-temporal compression of $32 \times 32 \times 8$ with 128 channels, resulting in a total compression ratio of 1:192 and a pixels-to-tokens ratio of 1:8192.

\begin{itemize}
    \item \textbf{Encoder:} The encoder is constructed using 3D Causal Convolutions to realize $32 \times 32 \times 8$ compression. Uniquely, to enable a unified encoding for both static images and video sequences, the first frame is encoded independently.
    \item \textbf{Decoder:} The decoder consists of several 3D convolution-based upsampling blocks and conditional convolutional residual blocks. The generation quality is modulated by the diffusion timestep and multi-layer noise as conditions.
\end{itemize}

\begin{figure}[t]
    \centering
    \includegraphics[width=0.7\linewidth]{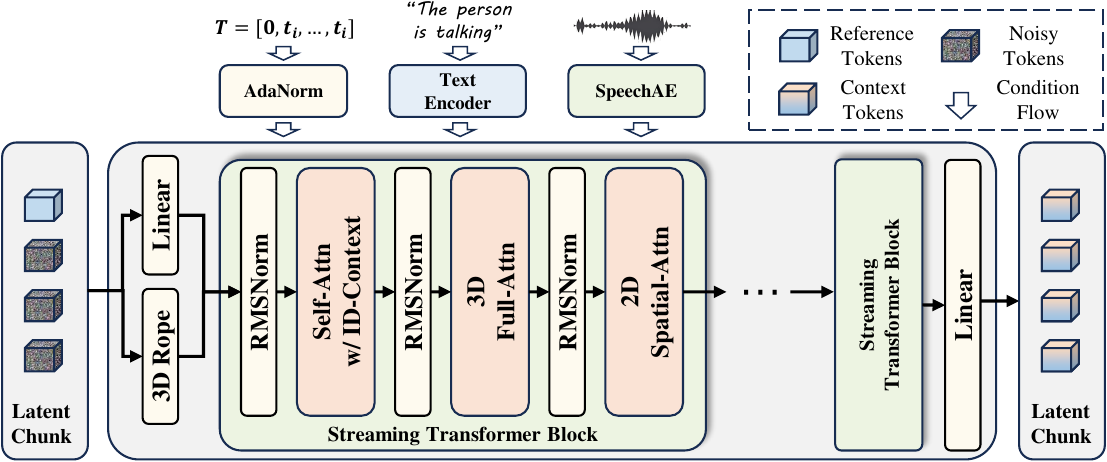}
    \caption{Detailed architecture of our Streaming A2V-DiT. Each transformer block encompass three attention mechanisms: self-attention with ID-Context Cache, 3D full-attention for text conditioning and 2D spatial-attention for speech conditioning.}
    \label{fig:a2vdit}
\end{figure}

\noindent\textbf{Detailed SpeechAE Architecture.}
Our SpeechAE architecture draws inspiration from READ~\cite{read}. Still, it distinguishes itself by utilizing multi-layer hierarchical pre-trained deep features as the raw input (refer to Sec.~\ref{sec:preprocess} for feature extraction details). The detailed architecture is illustrated in Fig.~\ref{fig:speechae}.

\begin{itemize}
    \item \textbf{Encoder-Decoder Structure:} The SpeechAE employs a fully convolutional encoder-decoder architecture. The SpeechAE Encoder comprises a linear projection block for feature dimension transformation and three convolutional downsampling blocks for the temporal compression of the hierarchical deep audio features. The Decoder utilizes causal 1D convolution-based residual blocks and upsampling modules to reconstruct the original input features from the compressed latent codes.
    
    \item \textbf{Audio-Visual Alignment:} A critical design aspect of the SpeechAE is ensuring precise alignment with the video latent space. To facilitate this, the first frame of the raw audio features is processed separately, mirroring the strategy employed for the video latent representation in our Temporal VAE.
    
    \item \textbf{Formulation:} The raw audio features $\boldsymbol{S}_{1:F} \in \mathbb{R}^{F\times H_w \times D_{\text{A}}}$ are processed by the SpeechAE Encoder to produce temporally compressed speech latent codes $\boldsymbol{E}\in\mathbb{R}^{f \times h_w\times d_{\text{A}}}$, where $h_w=2$, $d_{\text{A}}=2048$, and $f=(F-1)//8+1$ in our implementation, as formulated in Eq.~\ref{eqa:speechae} of Sec.~\ref{sec:idcontext}.
\end{itemize}

\noindent\textbf{Detailed Cross-Modal Diffusion Transformer (A2V-DiT).}
Our Streaming A2V-DiT model is built upon a Diffusion Transformer (DiT)~\cite{diffusiontransformer} architecture, as illustrated in Fig.~\ref{fig:a2vdit}. The input to the backbone is the initial noisy video latent after being patchified. The shape of the patchified video latent is $[h\times w \times f, d]$, where $h=16, w=16, f=13, d=128$ in our implementation. These patchified tokens are projected into a consistent latent space with a hidden dimensionality of 2048 via a linear layer. To encode positional information across both space and time, we apply the 3D Rotary Position Embeddings (RoPE)~\cite{rope1,rope2} to the patchified video tokens. The core design of the Streaming A2V-DiT consists of 28 transformer blocks, each encompass a self-attention module for historical information conditioning, a spatial audio cross-attention module for speech conditioning and a 3D full-attention module for text conditioning. The model generation is guided by several conditional inputs:
\begin{itemize}
    \item \textbf{History Condition:} Each transformer block in the Streaming A2V-DiT first integrates our proposed ID-Context Cache self-attention module to facilitate interaction between historical information and the current frame chunk. The ID-Context Cache maintains a KV cache length of 7 frames: the first frame is fixed as the reference image's KV value following the ID-Sink principle, while the subsequent 6 frames store the KV values of the preceding and current chunks. We employ a temporal sliding window mechanism where the KV cache of the preceding chunk is discarded to accommodate the KV pairs of the incoming chunk. The self-attention mechanism is formulated in Eqs.~\ref{eq:selfattn1}--\ref{eq:selfattn3}. 
    
    \item \textbf{Speech Condition:} The speech latent codes generated by the previously described SpeechAE Encoder serve as input to a spatial audio attention module. Notably, to strictly preserve the chronological alignment between audio and video, no caching mechanism is applied during this cross-modal attention step, formulated in Eq.~\ref{eq:speechattn} of Sec.~\ref{sec:idcontext}.
    
    \item \textbf{Text Condition:} Text embeddings are extracted from the text prompt and attention mask using a T5~\cite{t5} encoder and are subsequently fed into a 3D full-attention module. For training efficiency, we utilize a global text embedding pre-extracted offline via the T5 model, further details in Sec.~\ref{sec:trainingcfg}.
\end{itemize}

Additionally, the asynchronous noise timestep is encoded via an AdaLayerNorm~\cite{diffusiontransformer} module and incorporated into each attention layer to condition the noise prediction. Finally, the transformer output is projected back to the target video latent space. This design ensures effective audio-visual alignment while achieving semi-autoregressive streaming generation with high computational efficiency, forming a robust foundation for real-time talking head generation.

\section{Training and Inference Details}
\label{sec:traindetails}
This section supplements Sec.~\ref{sec:ASD} by providing a comprehensive exposition of the proposed Asynchronous Streaming Distillation (ASD) strategy, presenting a rigorous theoretical analysis that elucidates the underlying principles and mathematical rationale of our designed loss function. Furthermore, the chapter delineates the methodological framework and technical implementation specifics for model training and streaming inference utilizing ASD.

\subsection{Theoretical Analysis of the ASD Contrastive Objective}
\label{sec:proveasdcon}

In this section, we provide a rigorous theoretical derivation of the proposed Asynchronous Streaming Distillation (ASD) contrastive objective. We proceed from first principles: establishing the maximization of frame-wise Mutual Information (MI) as the requisite optimization target for temporal consistency, and systematically deriving the tractable contrastive loss function used in the proposed ASD of our framework.

\subsubsection{Problem Formulation: Spatio-Temporal Distribution Alignment}
The fundamental challenge in our framework is the information asymmetry between the non-streaming teacher ($\mathcal{T}$) and the streaming student ($\mathcal{S}$). The non-streaming teacher accesses the full temporal context during generation, whereas the student is strictly causal. To mitigate regression-to-the-mean artifacts~\cite{infonce} caused by MSE minimization under uncertainty, we reframe the contrastive distillation objective as distributional alignment between teacher and student models.

As demonstrated in Sec.~\ref{sec:ASD} of the main text, it is worth noticing that we treat the video not as a single high-dimensional vector, but as a sequence of random variables. Let the clean latent content of the $i$-th frame of $\boldsymbol{Z}(0)$ be $\boldsymbol{z}_i(0) \in \mathbb{R}^{D_{\text{frame}}}$, where $D_{\text{frame}} = h \cdot w \cdot d_\text{v}$, where $h$ and $w$ represents the latent height and width and $d_\text{v}$ represents the latent dim of the compressed video latent space, as described in Sec.~\ref{sec:idcontext}. We propose that a robust student model should maximize the statistical dependence between its generated frame distribution and the teacher's frame distribution at the corresponding timestamp. Formally, we aim to maximize the average Mutual Information (MI) over the sequence of length $f$, where $f$ represents the number of latent frames of $\boldsymbol{Z}(0)$ after compression, as described in Sec.~\ref{sec:idcontext}:
\begin{equation}
    \max_{\theta} \mathcal{J} = \frac{1}{f} \sum_{i=1}^{f} I(\boldsymbol{z}^\mathcal{T}_i(0); \boldsymbol{z}^\mathcal{S}_i(0))
    \label{eq:objective_z}
\end{equation}

\subsubsection{Equivalence of Velocity and Content Alignment}
\label{sec:proveasdcon1}

A practical obstacle is that Flow Matching models predict velocity fields $\boldsymbol{v}$ rather than clean latents $\boldsymbol{Z}(0)$ during training. We first prove that aligning the velocity frames is mathematically equivalent to aligning the clean content frames.

\begin{theorem}[Frame-wise Invariance of Mutual Information]
\label{thm:equivalence}
    Let $\boldsymbol{v}^\mathcal{T}_i$ and $\boldsymbol{v}^\mathcal{S}_i$ be the velocity feature maps for the $i$-th frame predicted by the teacher and student, respectively. Maximizing the mutual information between the reconstructed clean frames is equivalent to maximizing the mutual information between the predicted velocity frames:
    \begin{equation}
        I(\boldsymbol{z}^\mathcal{T}_i(0); \boldsymbol{z}^\mathcal{S}_i(0)) = I(\boldsymbol{v}^\mathcal{T}_i; \boldsymbol{v}^\mathcal{S}_i)
    \end{equation}
\end{theorem}

\begin{proof}
    Consider the conditional Flow Matching objective at a specific frame index $i$. The relationship between the predicted velocity $\boldsymbol{v}_i$, the sampled noise $\boldsymbol{\epsilon}_i$, and the reconstructed clean data $\boldsymbol{Z}_i(0)$ is governed by the linear equation:
    \begin{equation}
        \boldsymbol{v}_i = \boldsymbol{\epsilon}_i - \boldsymbol{z}_i(0)
    \end{equation}
    In our distillation framework, for any given training instance, both the teacher and student are conditioned on the identical noise sample $\boldsymbol{\epsilon}_i$ for that frame. We define the transformation function $g_{\boldsymbol{\epsilon}_i}(\cdot)$ as:
    \begin{equation}
        \boldsymbol{z}_i(0) = g_{\boldsymbol{\epsilon}_i}(\boldsymbol{v}_i) = \boldsymbol{\epsilon}_i - \boldsymbol{v}_i
    \end{equation}
    The function $g_{\boldsymbol{\epsilon}_i}: \mathbb{R}^{D_{\text{frame}}} \to \mathbb{R}^{D_{\text{frame}}}$ represents a translation followed by a reflection in the feature space. This defines a linear bijection (diffeomorphism) with a constant Jacobian determinant $|\det(J_{g})| = 1$.
    
    A fundamental property of Mutual Information is its invariance under smooth invertible transformations. Specifically, for random variables $X, Y$ and bijectors $\phi, \psi$, it holds that $I(X; Y) = I(\phi(X); \psi(Y))$. Applying this property yields:
    \begin{equation}
        I(\boldsymbol{z}^\mathcal{T}_i(0); \boldsymbol{z}^\mathcal{S}_i(0)) 
        = I(\boldsymbol{\epsilon}_i - \boldsymbol{v}^\mathcal{T}_i; \boldsymbol{\epsilon}_i - \boldsymbol{v}^\mathcal{S}_i) 
        = I(\boldsymbol{v}^\mathcal{T}_i; \boldsymbol{v}^\mathcal{S}_i)
    \end{equation}
    This theorem justifies defining the optimization target directly on the network output velocity frames.
\end{proof}

\subsubsection{Derivation of the Temporal Contrastive Bound}
Direct computation of the mutual information between student and teacher representations is generally intractable. We therefore relate the proposed temporal contrastive objective to a tractable variational lower bound~\cite{infonce2} by viewing distribution alignment as a \emph{temporal synchronization} problem. Concretely, for each student velocity frame $\boldsymbol{v}^\mathcal{S}_i$, the model must classify the correct synchronous teacher velocity frame $\boldsymbol{v}^\mathcal{T}_i$ (positive sample) against $f-1$ asynchronous teacher frames $\{\boldsymbol{v}^\mathcal{T}_j\}_{j\neq i}$ (negative samples) drawn from the same sequence.

\begin{theorem}[InfoNCE as a Temporal Mutual Information Bound]
\label{thm:infonce}
Let ${\mathit{\boldsymbol{v}^{\mathcal{S}}}}=\{\boldsymbol{v}^\mathcal{S}_i\}_{i=0}^f$ and ${\mathit{\boldsymbol{v}^{\mathcal{T}}}}=\{\boldsymbol{v}^\mathcal{T}_i\}_{i=0}^f$, where $\boldsymbol{v}^\mathcal{S}_i,\boldsymbol{v}^\mathcal{T}_i\in\mathbb{R}^{h\times w\times d_v}$. For each anchor frame $i$, the asynchronous teacher frames used as negatives are (approximately) drawn from the teacher marginal distribution $p(\boldsymbol{v}^\mathcal{T})$, implemented by uniformly sampling timestamps that are not synchronized with $i$ within the same sequence. Then minimizing the temporal contrastive loss $\mathcal{L}_{\text{CON}}$ maximizes a rigorous lower bound on the average mutual information between synchronous frames (excludes $i=0$ because the first frame serves as a shared reference image latent for both the teacher and student):
\begin{equation}
\frac{1}{f}\sum_{i=1}^{f} I(\boldsymbol{v}^\mathcal{S}_i;\boldsymbol{v}^\mathcal{T}_i)
\;\ge\; \log(f)\;-\;\mathcal{L}_{\text{CON}}.
\end{equation}
\end{theorem}

\begin{proof}
We formulate the distribution alignment of teacher and student as a temporal synchronization task. For each student velocity frame $\boldsymbol{v}^\mathcal{S}_i$, the model must classify the correct synchronous teacher velocity frame $\boldsymbol{v}^\mathcal{T}_i$ (positive sample) against $f-1$ asynchronous velocity frames $\{\boldsymbol{v}^\mathcal{T}_j\}_{j \neq i}$ (negative samples) from the same sequence. As demonstrated in Sec.~\ref{sec:ASD}, the proposed contrastive loss function of ASD is defined as:
\begin{equation}
  \mathcal{L}_{\text{CON}} = - \frac{1}{f} \sum_{i=1}^{f} \log \left( \frac{\exp \left( \mathrm{sim}(\boldsymbol{v}^\mathcal{S}_i, \boldsymbol{v}^\mathcal{T}_i) / \tau \right)}{ \sum\limits_{j=1}^{f} \exp \left( \mathrm{sim}(\boldsymbol{v}^\mathcal{S}_i, \boldsymbol{v}^\mathcal{T}_j) / \tau \right) } \right).
\label{eq:con_def}
\end{equation}

Define the score function as follows:
\begin{equation}
f_\theta(\boldsymbol{v}^\mathcal{S}_i,\boldsymbol{v}^\mathcal{T}_j)
\triangleq
\mathrm{sim}(\boldsymbol{v}^\mathcal{S}_i,\boldsymbol{v}^\mathcal{T}_j)/\tau,
\end{equation}
and the softmax distribution over teacher indices induced by this score:
\begin{equation}
p_\theta\!\left(j \,\middle|\, \boldsymbol{v}^\mathcal{S}_i, \{\boldsymbol{v}^\mathcal{T}_k\}_{k=1}^f\right)
\triangleq
\frac{\exp\!\left(f_\theta(\boldsymbol{v}^\mathcal{S}_i,\boldsymbol{v}^\mathcal{T}_j)\right)}
{\sum_{k=1}^{f}\exp\!\left(f_\theta(\boldsymbol{v}^\mathcal{S}_i,\boldsymbol{v}^\mathcal{T}_k)\right)}.
\label{eq:softmax}
\end{equation}
For each anchor $\boldsymbol{v}^\mathcal{S}_i$, the logarithm in \eqref{eq:con_def} is precisely the log-likelihood assigned by \eqref{eq:softmax} to the synchronized index $j=i$. Thus, $\mathcal{L}_{\text{CON}}$ is the average negative log-likelihood of classifying the synchronized teacher frame among $f$ candidates. To make the true label explicit, introduce a single random index $\kappa\sim \mathrm{Unif}\{1,\dots,f\}$ denoting the position of the synchronized teacher frame among the $f$ candidates. Under the standard InfoNCE~\cite{infonce} synchronization sampling scheme, the synchronized pair $(\boldsymbol{v}^\mathcal{S}_\kappa,\boldsymbol{v}^\mathcal{T}_\kappa)$ is drawn from the joint distribution of synchronous frames $p(\boldsymbol{v}^\mathcal{S},\boldsymbol{v}^\mathcal{T})$, while the remaining candidates $\{\boldsymbol{v}^\mathcal{T}_j\}_{j\neq \kappa}$ are drawn independently from the marginal $p(\boldsymbol{v}^\mathcal{T})$. This defines the joint distribution
\begin{equation}
q\!\left(\kappa,\boldsymbol{v}^\mathcal{S}_\kappa,\{\boldsymbol{v}^\mathcal{T}_j\}_{j=1}^f\right)
\triangleq
\frac{1}{f}\,p\!\left(\boldsymbol{v}^\mathcal{S}_\kappa,\boldsymbol{v}^\mathcal{T}_\kappa\right)\prod_{j\neq \kappa}p\!\left(\boldsymbol{v}^\mathcal{T}_j\right).
\label{eq:q}
\end{equation}
Because the denominator in \eqref{eq:softmax} is symmetric over indices, randomizing the positive location via $\kappa$ does not change the objective value; it only provides a convenient probabilistic representation for analysis. For later convenience, define the marginal (mixture) distribution obtained by summing out the index:
\begin{equation}
\tilde q\!\left(\boldsymbol{v}^\mathcal{S},\{\boldsymbol{v}^\mathcal{T}_j\}_{j=1}^f\right)
\triangleq
\sum_{\kappa=1}^{f}
q\!\left(\kappa,\boldsymbol{v}^\mathcal{S},\{\boldsymbol{v}^\mathcal{T}_j\}_{j=1}^f\right)
=
\frac{1}{f}\sum_{k=1}^{f}
p(\boldsymbol{v}^\mathcal{S},\boldsymbol{v}^\mathcal{T}_k)\prod_{j\neq k}p(\boldsymbol{v}^\mathcal{T}_j).
\label{eq:q_tilde}
\end{equation}

Consider the expected negative log-likelihood of predicting the synchronized index under \eqref{eq:q}:
\begin{equation}
\mathcal{J}_\theta
\triangleq
\mathbb{E}_{q}\!\left[-\log p_\theta\!\left(\kappa \,\middle|\, \boldsymbol{v}^\mathcal{S}_\kappa,\{\boldsymbol{v}^\mathcal{T}_j\}_{j=1}^f\right)\right].
\label{eq:J}
\end{equation}
By construction, $\mathcal{J}_\theta$ is the population analogue of the empirical objective in \eqref{eq:con_def} (averaged over anchors); hence $\mathcal{J}_\theta$ corresponds to $\mathcal{L}_{\text{CON}}$ under the sampling scheme in the theorem. Since \eqref{eq:J} is a cross-entropy between the true posterior $q(\kappa\mid \boldsymbol{v}^\mathcal{S}_\kappa,\{\boldsymbol{v}^\mathcal{T}_j\}_{j=1}^f)$ and the model distribution $p_\theta(\cdot\mid \boldsymbol{v}^\mathcal{S}_\kappa,\{\boldsymbol{v}^\mathcal{T}_j\}_{j=1}^f)$, the non-negativity of KL divergence implies:
\begin{align}
\mathcal{J}_\theta
&\ge
\mathbb{E}_{q}\!\left[-\log q\!\left(\kappa \,\middle|\, \boldsymbol{v}^\mathcal{S}_\kappa,\{\boldsymbol{v}^\mathcal{T}_j\}_{j=1}^f\right)\right]
=
H_q\!\left(\kappa \,\middle|\, \boldsymbol{v}^\mathcal{S}_\kappa,\{\boldsymbol{v}^\mathcal{T}_j\}_{j=1}^f\right).
\label{eq:CE_ge_H}
\end{align}
Because $\kappa$ is uniform under \eqref{eq:q}, $H(\kappa)=\log f$, and therefore
\begin{equation}
H_q\!\left(\kappa \,\middle|\, \boldsymbol{v}^\mathcal{S}_\kappa,\{\boldsymbol{v}^\mathcal{T}_j\}_{j=1}^f\right)
=
\log f
-
I_q\!\left(\kappa;\boldsymbol{v}^\mathcal{S}_\kappa,\{\boldsymbol{v}^\mathcal{T}_j\}_{j=1}^f\right).
\label{eq:H_to_I}
\end{equation}
Thus, lower bounding $-\mathcal{J}_\theta$ reduces to upper bounding the index mutual information
$I_q(\kappa;\boldsymbol{v}^\mathcal{S}_\kappa,\{\boldsymbol{v}^\mathcal{T}_j\}_{j=1}^f)$.

Applying Bayes' rule to \eqref{eq:q} yields the true posterior over indices:
\begin{equation}
q\!\left(\kappa=i \,\middle|\, \boldsymbol{v}^\mathcal{S}_\kappa,\{\boldsymbol{v}^\mathcal{T}_j\}_{j=1}^f\right)
=
\frac{\frac{p(\boldsymbol{v}^\mathcal{T}_i\mid \boldsymbol{v}^\mathcal{S}_\kappa)}{p(\boldsymbol{v}^\mathcal{T}_i)}}
{\sum_{k=1}^{f}\frac{p(\boldsymbol{v}^\mathcal{T}_k\mid \boldsymbol{v}^\mathcal{S}_\kappa)}{p(\boldsymbol{v}^\mathcal{T}_k)}}.
\label{eq:true_posterior}
\end{equation}
Using $q(\kappa)=1/f$ and \eqref{eq:true_posterior} to obtain the decomposition as follows:
\begin{align}
I_q\!\left(\kappa;\boldsymbol{v}^\mathcal{S}_\kappa,\{\boldsymbol{v}^\mathcal{T}_j\}_{j=1}^f\right)
&=
\mathbb{E}_{q}\!\left[\log\frac{p(\boldsymbol{v}^\mathcal{T}_\kappa\mid \boldsymbol{v}^\mathcal{S}_\kappa)}{p(\boldsymbol{v}^\mathcal{T}_\kappa)}\right]
-
\mathbb{E}_{q}\!\left[\log\!\left(\frac{1}{f}\sum_{k=1}^{f}\frac{p(\boldsymbol{v}^\mathcal{T}_k\mid \boldsymbol{v}^\mathcal{S}_\kappa)}{p(\boldsymbol{v}^\mathcal{T}_k)}\right)\right].
\label{eq:Iq_decompose}
\end{align}
The first expectation equals the mutual information of synchronized frames because $(\boldsymbol{v}^\mathcal{S}_\kappa,\boldsymbol{v}^\mathcal{T}_\kappa)\sim p(\boldsymbol{v}^\mathcal{S},\boldsymbol{v}^\mathcal{T})$ under \eqref{eq:q}:
\begin{equation}
\mathbb{E}_{q}\!\left[\log\frac{p(\boldsymbol{v}^\mathcal{T}_\kappa\mid \boldsymbol{v}^\mathcal{S}_\kappa)}{p(\boldsymbol{v}^\mathcal{T}_\kappa)}\right]
=
I\!\left(\boldsymbol{v}^\mathcal{S}_\kappa;\boldsymbol{v}^\mathcal{T}_\kappa\right).
\label{eq:first_term}
\end{equation}
To handle the second term, define the all-marginal reference distribution
\begin{equation}
r\!\left(\boldsymbol{v}^\mathcal{S}_\kappa,\{\boldsymbol{v}^\mathcal{T}_j\}_{j=1}^f\right)
\triangleq
p(\boldsymbol{v}^\mathcal{S}_\kappa)\prod_{j=1}^{f}p(\boldsymbol{v}^\mathcal{T}_j).
\end{equation}
A direct calculation from \eqref{eq:q_tilde} yields
\begin{equation}
\frac{\tilde q\!\left(\boldsymbol{v}^\mathcal{S}_\kappa,\{\boldsymbol{v}^\mathcal{T}_j\}_{j=1}^f\right)}
{r\!\left(\boldsymbol{v}^\mathcal{S}_\kappa,\{\boldsymbol{v}^\mathcal{T}_j\}_{j=1}^f\right)}
=
\frac{1}{f}\sum_{k=1}^{f}\frac{p(\boldsymbol{v}^\mathcal{T}_k\mid \boldsymbol{v}^\mathcal{S}_\kappa)}{p(\boldsymbol{v}^\mathcal{T}_k)}.
\end{equation}
Therefore, the second expectation in \eqref{eq:Iq_decompose} equals a KL divergence and is non-negative:
\begin{equation}
\mathbb{E}_{\tilde q}\!\left[\log\!\left(\frac{1}{f}\sum_{k=1}^{f}\frac{p(\boldsymbol{v}^\mathcal{T}_k\mid \boldsymbol{v}^\mathcal{S}_\kappa)}{p(\boldsymbol{v}^\mathcal{T}_k)}\right)\right]
=
\mathrm{KL}\!\left(\tilde q\;\|\;r\right)
\ge 0.
\label{eq:KL}
\end{equation}
Combining \eqref{eq:Iq_decompose}--\eqref{eq:KL} yields
\begin{equation}
I_q\!\left(\kappa;\boldsymbol{v}^\mathcal{S}_\kappa,\{\boldsymbol{v}^\mathcal{T}_j\}_{j=1}^f\right)
\le
I\!\left(\boldsymbol{v}^\mathcal{S}_\kappa;\boldsymbol{v}^\mathcal{T}_\kappa\right).
\end{equation}
Substituting this inequality into \eqref{eq:H_to_I} and then into \eqref{eq:CE_ge_H} gives
\begin{equation}
\mathcal{J}_\theta
\ge
\log f - I\!\left(\boldsymbol{v}^\mathcal{S}_\kappa;\boldsymbol{v}^\mathcal{T}_\kappa\right),
\end{equation}
which rearranges to
\begin{equation}
I\!\left(\boldsymbol{v}^\mathcal{S}_\kappa;\boldsymbol{v}^\mathcal{T}_\kappa\right)
\ge
\log f - \mathcal{J}_\theta.
\label{eq:MI_bound}
\end{equation}

Finally, since $\kappa$ is uniform on $\{1,\dots,f\}$, the left-hand side of \eqref{eq:MI_bound} equals $\frac{1}{f}\sum_{i=1}^{f} I(\boldsymbol{v}^\mathcal{S}_i;\boldsymbol{v}^\mathcal{T}_i)$. Moreover, $\mathcal{J}_\theta$ is the population counterpart of the empirical objective defining $\mathcal{L}_{\text{CON}}$. Therefore, the proof is completed:
\begin{equation}
\frac{1}{f}\sum_{i=1}^{f} I(\boldsymbol{v}^\mathcal{S}_i;\boldsymbol{v}^\mathcal{T}_i)
\ge
\log f - \mathcal{L}_{\text{CON}},
\end{equation}

\end{proof}

In summary, minimizing $\mathcal{L}_{\text{CON}}$ maximizes a rigorous lower bound on the frame-wise average mutual information between ${\mathit{\boldsymbol{v}^{\mathcal{S}}}}$ and ${\mathit{\boldsymbol{v}^{\mathcal{T}}}}$. By contrasting synchronized pairs $(\boldsymbol{v}^\mathcal{S}_i,\boldsymbol{v}^\mathcal{T}_i)$ against temporally asynchronous negatives $(\boldsymbol{v}^\mathcal{S}_i,\boldsymbol{v}^\mathcal{T}_j)$ with $j\neq i$, the student is encouraged to preserve the teacher's temporal correspondence and motion evolution, consistent with the analysis in Sec.~\ref{sec:proveasdcon1}.

\subsection{Theoretical Analysis of the ASD Smoothness Objective}
\label{sec:proveasdsmo}
In this section, we provide a rigorous theoretical justification for the proposed Second-Order Smoothness Matching Objective ($\mathcal{L}_{\text{SMO}}$). We formulate the analysis strictly within the discrete domain, treating the teacher and student outputs as finite sequences $\boldsymbol{v}^{\mathcal{T}} = [\boldsymbol{v}^\mathcal{T}_0, \dots, \boldsymbol{v}^\mathcal{T}_f]$ and $\boldsymbol{v}^{\mathcal{S}} = [\boldsymbol{v}^\mathcal{S}_0, \dots, \boldsymbol{v}^\mathcal{S}_f]$.
We demonstrate that $\mathcal{L}_{\text{SMO}}$ satisfies three critical properties: (1) strict isometry to content smoothness; (2) general high-pass filtering capability that suppresses discrete high-frequency jitter; and (3) orthogonality to the permutation-invariant nature of the contrastive loss.

\subsubsection{Equivalence of Discrete Velocity and Data Smoothness}
We first prove that optimizing the discrete second-order difference in the velocity space is mathematically identical to optimizing it in the clean latent space.

\begin{lemma}[Isometry of Discrete Differences]
\label{lemma:isometry}
    Let $\Delta^2$ denote the discrete second-order difference operator defined as $\Delta^2 \boldsymbol{x}_i = \boldsymbol{x}_{i+1} - 2\boldsymbol{x}_i + \boldsymbol{x}_{i-1}$. Under the Flow Matching framework with shared noise conditioning, the loss in the velocity domain $\mathcal{L}_{\text{SMO}}^{\boldsymbol{v}}$ is equivalent to the loss in the clean latent domain $\mathcal{L}_{\text{SMO}}^{\boldsymbol{Z}}$:
    \begin{equation}
        \frac{1}{f-1}\sum_{i=1}^{f-1} ||\Delta^2 \boldsymbol{v}^\mathcal{S}_i - \Delta^2 \boldsymbol{v}^\mathcal{T}_i||^2 \equiv \frac{1}{f-1}\sum_{i=1}^{f-1} ||\Delta^2 \boldsymbol{Z}^\mathcal{S}_i(0) - \Delta^2 \boldsymbol{Z}^\mathcal{T}_i(0)||^2
    \end{equation}
\end{lemma}

\begin{proof}
    For any discrete time step $i$, the Flow Matching formulation defines the linear relationship: $\boldsymbol{v}_i = \boldsymbol{\epsilon}_i - \boldsymbol{Z}_i(0)$.
    During the distillation process, the Teacher ($\mathcal{T}$) and Student ($\mathcal{S}$) are conditioned on the identical discrete noise sequence $\boldsymbol{\epsilon} = [\boldsymbol{\epsilon}_0, \dots, \boldsymbol{\epsilon}_f]$.
    We expand the difference term in the velocity loss:
    \begin{align}
        \Delta^2 \boldsymbol{v}^\mathcal{S}_i - \Delta^2 \boldsymbol{v}^\mathcal{T}_i &= \Delta^2 (\boldsymbol{\epsilon}_i - \boldsymbol{Z}^\mathcal{S}_i(0)) - \Delta^2 (\boldsymbol{\epsilon}_i - \boldsymbol{Z}^\mathcal{T}_i(0))
    \end{align}
    Since $\Delta^2$ is a linear operator on the sequence space, we can rearrange the terms:
    \begin{align}
        &= (\Delta^2 \boldsymbol{\epsilon}_i - \Delta^2 \boldsymbol{\epsilon}_i) - (\Delta^2 \boldsymbol{Z}^\mathcal{S}_i(0) - \Delta^2 \boldsymbol{Z}^\mathcal{T}_i(0)) \\
        &= -(\Delta^2 \boldsymbol{Z}^\mathcal{S}_i(0) - \Delta^2 \boldsymbol{Z}^\mathcal{T}_i(0))
    \end{align}
    Squaring the norm eliminates the negative sign:
    \begin{equation}
        ||\Delta^2 \boldsymbol{v}^\mathcal{S}_i - \Delta^2 \boldsymbol{v}^\mathcal{T}_i||^2 = ||\Delta^2 \boldsymbol{Z}^\mathcal{S}_i(0) - \Delta^2 \boldsymbol{Z}^\mathcal{T}_i(0)||^2
    \end{equation}
    Summing over all $i$ proves the equivalence. Thus, minimizing the discrete jerk of the velocity sequence directly imposes smoothness on the generated content sequence.
\end{proof}

\subsubsection{Suppression of Discrete High-Frequency Jitter}
To rigorously prove the capacity of $\mathcal{L}_{\text{SMO}}$ to suppress high-frequency artifacts generally (beyond specific error models), we employ frequency domain analysis via the Z-transform.

\begin{theorem}[Spectral Analysis of the Smoothness Objective]
\label{thm:jitter}
    The smoothness loss $\mathcal{L}_{\text{SMO}}$ functions as a high-pass filter in the frequency domain. For any generalized error sequence with normalized angular frequency $\omega \in [0, \pi]$, the loss applies a spectral gain of $16 \sin^4(\omega/2)$. This strictly suppresses low-frequency components while aggressively penalizing high-frequency jitter, reaching a maximum penalty at the Nyquist frequency ($\omega=\pi$).
\end{theorem}

\begin{proof}
    Let $\boldsymbol{e}_n = \boldsymbol{v}^\mathcal{S}_n - \boldsymbol{v}^\mathcal{T}_n$ be the discrete error sequence between student and teacher. The objective $\mathcal{L}_{\text{SMO}}$ minimizes the energy of the second-order difference signal defined by $y_n = \Delta^2 \boldsymbol{e}_n = \boldsymbol{e}_{n+1} - 2\boldsymbol{e}_n + \boldsymbol{e}_{n-1}$.
    
    We analyze this operation in the Z-domain. The transfer function $H(z)$ corresponding to the difference operator is:
    \begin{equation}
        H(z) = z^1 - 2 + z^{-1}
    \end{equation}
    To obtain the Frequency Response, we evaluate $H(z)$ on the unit circle by substituting $z = e^{j\omega}$:
    \begin{equation}
        H(e^{j\omega}) = e^{j\omega} - 2 + e^{-j\omega} = 2\cos(\omega) - 2
    \end{equation}
    The effective penalty applied by the loss function is proportional to the squared magnitude of the frequency response (Power Spectral Density gain):
    \begin{equation}
        |H(e^{j\omega})|^2 = (2\cos(\omega) - 2)^2 = 4(1 - \cos(\omega))^2
    \end{equation}
    Using the trigonometric identity $1 - \cos(\omega) = 2\sin^2(\omega/2)$, we derive the final spectral gain:
    \begin{equation}
        |H(e^{j\omega})|^2 = 16 \sin^4\left(\frac{\omega}{2}\right)
    \end{equation}
    This spectral analysis reveals the general filtering characteristics of $\mathcal{L}_{\text{SMO}}$:
    \begin{enumerate}
        \item Low-Frequency Suppression: As $\omega \to 0$, the gain approaches 0.
        \item Monotonic High-Frequency Penalization: The gain increases strictly with frequency.
        \item Nyquist Limit: At the Nyquist frequency $\omega = \pi$, the gain reaches its maximum of $16 \sin^4(\pi/2) = 16$.
    \end{enumerate}
    This confirms that the loss acts as a generic high-pass filter. The specific case of alternating noise discussed in subsequent sections corresponds to the worst-case spectral response at $\omega=\pi$.
\end{proof}

\subsection{Theoretical Analysis of the Complementarity of ASD Objectives}
\label{sec:suppanalysis}

In this section, we provide a theoretical analysis to justify the necessity of combining the temporal contrastive loss $\mathcal{L}_{\text{CON}}$ with the smoothness matching objective $\mathcal{L}_{\text{SMO}}$. We demonstrate that they impose complementary constraints on different derivative orders of the video sequence: $\mathcal{L}_{\text{CON}}$ enforces a 0-th order spatial alignment, whereas $\mathcal{L}_{\text{SMO}}$ provides a 2-nd order dynamic regularization. To formalize this analysis, we introduce a specific perturbation model representing high-frequency jitter, which corresponds to the Nyquist mode ($\omega=\pi$) analyzed generally in Theorem \ref{thm:jitter}.

\begin{theorem}[Orthogonal Sensitivity to High-Frequency Noise]
\label{thm:complementarity}
    Consider a scenario where the student's output $\boldsymbol{v}^\mathcal{S}_i$ is a perturbed version of the teacher's output $\boldsymbol{v}^\mathcal{T}_i$, corrupted by a high-frequency, low-amplitude oscillation error. This relationship can be modeled as:
    \begin{equation}
        \boldsymbol{v}^\mathcal{S}_i = \boldsymbol{v}^\mathcal{T}_i + \boldsymbol{\delta}_i
    \end{equation}
    where the error term is defined as alternating noise $\boldsymbol{\delta}_i = (-1)^i \epsilon \boldsymbol{u}$, with a unit direction vector $||\boldsymbol{u}||=1$, orthogonal to the feature vector ($\boldsymbol{u} \perp \boldsymbol{v}^\mathcal{T}_i$), and a small magnitude $\epsilon \to 0$. Under this formulation:
    \begin{enumerate}
        \item \textbf{The disappearance gradient of InfoNCE:} The contrastive loss $\mathcal{L}_{\text{CON}}$ exhibits a vanishing first-order gradient with respect to the noise amplitude $\epsilon$, which scales quadratically ($O(\epsilon^2)$). This implies the loss surface is locally flat to infinitesimal jitter.
        \item \textbf{Spectral Amplification of SMO:} The smoothness loss $\mathcal{L}_{\text{SMO}}$ amplifies the high-frequency error components, scaling quadratically with a factor of 16 relative to the base error magnitude ($16\epsilon^2$).
    \end{enumerate}
\end{theorem}

\begin{proof}
    \textbf{1. Analysis of $\mathcal{L}_{\text{CON}}$ (0-th Order Constraint)}
    
    We first analyze the behavior of the contrastive loss. The core mechanism of $\mathcal{L}_{\text{CON}}$ relies on the cosine similarity between the student and teacher features. Let $\mathrm{sim}(\boldsymbol{a}, \boldsymbol{b}) = \frac{\boldsymbol{a}^T \boldsymbol{b}}{||\boldsymbol{a}|| ||\boldsymbol{b}||}$. Substituting the perturbed student state $\boldsymbol{v}^\mathcal{S}_i = \boldsymbol{v}^\mathcal{T}_i + \boldsymbol{\delta}_i$ into the similarity function:
    \begin{equation}
        \mathrm{sim}(\boldsymbol{v}^\mathcal{S}_i, \boldsymbol{v}^\mathcal{T}_i) = \frac{(\boldsymbol{v}^\mathcal{T}_i + \boldsymbol{\delta}_i)^T \boldsymbol{v}^\mathcal{T}_i}{||\boldsymbol{v}^\mathcal{T}_i + \boldsymbol{\delta}_i|| ||\boldsymbol{v}^\mathcal{T}_i||}
    \end{equation}
    Assuming $||\boldsymbol{v}^\mathcal{T}_i|| = 1$ for simplification (without loss of generality due to normalization layers) and orthogonality $\boldsymbol{\delta}_i \perp \boldsymbol{v}^\mathcal{T}_i$, the term simplifies via Taylor expansion for small $\epsilon$:
    \begin{equation}
        \mathrm{sim}(\boldsymbol{v}^\mathcal{S}_i, \boldsymbol{v}^\mathcal{T}_i) = \frac{1}{\sqrt{1+\epsilon^2}} \approx 1 - \frac{1}{2}\epsilon^2
    \end{equation}
    Now, we examine the sensitivity of the InfoNCE loss $\mathcal{L}_{\text{CON}}$. The loss for the $i$-th frame is defined as:
    \begin{equation}
        \mathcal{L}_i = - \log \frac{e^{s_{i,i}/\tau}}{e^{s_{i,i}/\tau} + \sum_{j \neq i} e^{s_{i,j}/\tau}}
    \end{equation}
    where $s_{i,i} = \mathrm{sim}(\boldsymbol{v}^\mathcal{S}_i, \boldsymbol{v}^\mathcal{T}_i)$ is the positive logit. 
    By the chain rule, the variation in the loss depends on the derivative of the similarity score with respect to the perturbation $\epsilon$. Crucially, at the unperturbed limit ($\epsilon \to 0$), this derivative vanishes:
    \begin{equation}
        \frac{\partial s_{i,i}}{\partial \epsilon} \approx \frac{\partial}{\partial \epsilon} \left( 1 - \frac{1}{2}\epsilon^2 \right) = -\epsilon \xrightarrow{\epsilon \to 0} 0
    \end{equation}
    Consequently, the first-order term in the Taylor expansion of the loss function is zero. The leading non-zero term for the contrastive loss variation is strictly quadratic:
    \begin{equation}
        \Delta \mathcal{L}_{\text{CON}} \approx \left| \frac{\partial \mathcal{L}}{\partial s_{i,i}} \right| \cdot \frac{1}{2}\epsilon^2 = O(\epsilon^2)
    \end{equation}
    This mathematical result indicates that the purely contrastive loss is perfectly insensitive (locally flat) to infinitesimal high-frequency orthogonal jitter. It provides negligible restoring gradients to suppress small-magnitude oscillations.

    \textbf{2. Analysis of $\mathcal{L}_{\text{SMO}}$ (2-nd Order Constraint)}
    
    In contrast, the smoothness objective $\mathcal{L}_{\text{SMO}}$ minimizes the squared difference of the discrete Laplacian operator $\Delta^2$. We proceed by calculating the divergence in the second-order variation:
    \begin{equation}
        \Delta^2 \boldsymbol{v}^\mathcal{S}_i - \Delta^2 \boldsymbol{v}^\mathcal{T}_i = \Delta^2 (\boldsymbol{v}^\mathcal{T}_i + \boldsymbol{\delta}_i) - \Delta^2 \boldsymbol{v}^\mathcal{T}_i = \Delta^2 \boldsymbol{\delta}_i
    \end{equation}
    We explicitly demonstrate the sensitivity of this term to the specific high-frequency noise pattern $\boldsymbol{\delta}_i = (-1)^i \epsilon \boldsymbol{u}$, which corresponds to the worst-case spectral response ($\omega=\pi$) derived in Theorem \ref{thm:jitter}. Substituting into the discrete difference formula:
    \begin{align}
        \Delta^2 \boldsymbol{\delta}_i &= \boldsymbol{\delta}_{i+1} - 2\boldsymbol{\delta}_i + \boldsymbol{\delta}_{i-1} \\
        &= (-1)^{i+1}\epsilon \boldsymbol{u} - 2(-1)^i\epsilon \boldsymbol{u} + (-1)^{i-1}\epsilon \boldsymbol{u}
    \end{align}
    Factoring out the common terms, we obtain:
    \begin{equation}
        \Delta^2 \boldsymbol{\delta}_i = \epsilon \boldsymbol{u} \left[ -(-1)^i - 2(-1)^i - (-1)^i \right] = -4 (-1)^i \epsilon \boldsymbol{u}
    \end{equation}
    Finally, substituting this result back into the loss function yields the scaling factor:
    \begin{equation}
        \mathcal{L}_{\text{SMO}} = \frac{1}{f-1}\sum ||-4 (-1)^i \epsilon \boldsymbol{u}||^2 = 16 \epsilon^2
    \end{equation}
    Comparing the two objectives: while both technically scale quadratically with $\epsilon$, $\mathcal{L}_{\text{CON}}$ suffers from a vanishing gradient effect governed by the soft probability distribution, whereas $\mathcal{L}_{\text{SMO}}$ acts as a spectral amplifier with a 16-fold penalty factor relative to the base error energy. The theoretical analysis presented above confirms that the two objectives of ASD are mathematically complementary:
    \begin{itemize}
        \item $\mathcal{L}_{\text{CON}}$ functions as a position anchor, ensuring that the student's output is semantically correct and synchronous at each timestamp $i$ (enforcing 0-th order location fidelity).
        \item $\mathcal{L}_{\text{SMO}}$ functions as a trajectory stabilizer, strictly penalizing inter-frame derivative anomalies and filtering out the high-frequency jitter that $\mathcal{L}_{\text{CON}}$ is locally insensitive to (enforcing 2-nd order dynamic fidelity).
    \end{itemize}
    Jointly, these two objectives ensure that the generated video streams of the streaming student model are not only semantically aligned with the non-streaming teacher model but also kinematically cohesive and stable.
\end{proof}

\subsection{Model Initialization}
To leverage robust visual generative priors and accelerate convergence, we initialize the core self-attention and 3D full-attention modules of our A2V-DiT backbone using the pre-trained weights from LTX-VIDEO~\cite{ltxvideo}. A critical architectural modification is the integration of the ID-Context Cache mechanism within the self-attention blocks. To prevent the abrupt disruption of pre-trained feature distributions, we employ a zero-initialization strategy for the ID-Context Cache. To equip the model with the novel capability of audio-driven talking head generation absent in the foundation pre-trained model, the newly introduced SpeechAE and 2D Audio Spatial Attention modules are initialized randomly. These parameters are subsequently optimized to model complex audio-visual cross-modal correlations during the talking head generation training.

\subsection{Device Information}

All experimental evaluations, including model training and inference, are implemented using the PyTorch framework with CUDA 12.6 and executed on NVIDIA A100 GPUs. Regarding computational resource consumption, the training phase incurred a peak video memory (VRAM) footprint of approximately 46GB with a batch size of 1, increasing to 70GB with a batch size of 2. During the inference phase, generating a 4.8-second video clip requires approximately 11GB of VRAM, demonstrating the feasibility of deployment. Furthermore, the cumulative computational budget for the entire pipeline, including both the optimization of the non-streaming teacher backbone and the ASD-guided streaming student distillation, amounts to approximately 400 GPU hours.

\subsection{Supplementary Training Details}
\label{sec:trainingcfg}

This section serves as a comprehensive supplement to Sec.~\ref{sec:ASD} of the main text, providing a granular exposition of the training configuration and procedural pipeline proposed for the REST framework. The training protocol is methodologically bifurcated into three distinct phases: (1) Non-Streaming Teacher Training, (2) Streaming Student Training, and (3) Asynchronous Streaming Distillation (ASD). In the following subsections, we delineate the specific mechanisms and hyperparameters governing each phase.

\subsubsection{Non-Streaming Teacher Training}

\label{sec:teacher_pretraining}

\noindent\textbf{Training Setup and Optimization Strategy.}
As outlined in Sec.~\ref{sec:ASD}, our training methodology for the non-streaming teacher model follows a coarse-to-fine two-stage optimization process. The temporal dimension of the training samples is fixed at $F=97$ frames in the pixel space, which corresponds to a sequence length of $f=13$ in the compressed latent space. We first pre-train the SpeechAE module to ensure robust audio feature extraction. This stage utilizes a learning rate of $\eta_1 = 1\times10^{-4}$ following the training strategy described in READ~\cite{read}. Subsequently, we train the main A2V-DiT backbone without ID-Context Cache. During this phase, the learning rate is decayed to $\eta_2 = 1\times10^{-5}$, and the pre-trained SpeechAE module is jointly fine-tuned with the generative backbone to ensure optimal audio-visual alignment.

\noindent\textbf{Chunk-wise Asynchronous Noise Scheduler (CANS).}
A core theoretical contribution of our training paradigm of the non-streaming teacher model is the simulation of streaming noise dynamics within a non-streaming environment. To achieve this, we employ a Chunk-wise Asynchronous Noise Scheduler (CANS). Firstly, we virtually partition the global latent sequence $\mathbf{Z}(0)$ into $K=4$ segments. The initial segment $\mathbf{z}_1$ comprises 4 latent frames (including one reference frame), while subsequent segments $\mathbf{z}_{i}$ ($i>1$) contain 3 frames each.

To simulate the accumulation of generation latency and effectively guide the training of the subsequent streaming student model, we impose an asynchronous noise distribution where earlier chunks are exposed to lower noise levels than later chunks. Specifically, we define a global noise schedule $\mathcal{N}_{\text{noise}}$ containing 1000 discrete timesteps sampled from a shifted-logit-normal distribution. For each virtual segment $i$, we sample a distinct timestep index $n_i$ from mutually exclusive intervals to strictly satisfy the monotonicity constraint. This process is formally expressed as:
\begin{equation}
    n_i \in \big[\:\ 250\,(i-1), \, 250\,i \: \big), \quad \forall \:i \in \{\,1,\, 2,\, 3,\, 4\,\}.
\end{equation}
The corresponding noise timestep for the $i$-th chunk is then assigned as $t_i = \mathcal{N}_{\text{noise}}[n_i]$. This configuration ensures that earlier segments (with lower $t$) preserve more structural information, thereby serving as a reliable context motion prior for the denoising process of subsequent noisy chunks.

\noindent\textbf{Dropout Regularization Strategies.}
To enhance model robustness and disentangle condition dependencies, we integrate three distinct dropout strategies during the forward diffusion process:
\begin{itemize}
    \item \textbf{Identity Dropout}: With a probability $p_{\text{id}}=0.1$, the reference image feature is zeroized ($\mathbf{I}_{\text{ref}} \to \mathbf{0}$). This forces the model to learn facial dynamics from the audio signal rather than relying solely on the identity reference for reconstruction. Improving the generalized generation capability of the teacher model.
    
    \item \textbf{Audio Dropout}: The audio condition is dropped with $p_{\text{aud}}=0.1$, reinforcing the effect of the speech signal through conditional guidance mechanisms and preventing overfitting to specific audio artifacts.
    
    \item \textbf{Motion Prior Dropout}: We introduce a motion prior dropout to the CANS forward noising process. With a probability of $p_{\text{mot}}=0.5$, we apply the asynchronous noise ($t_1 < t_2 < t_3 < t_4$) described above, simulating a conditional generation process guided by the valid motion prior of the preceding chunk. Conversely, for the remaining 50\%, we apply synchronous noise of uniform intensity ($t_1 = t_2 = t_3 = t_4$) across all frames. This simulates an unconditional generation scenario where no reliable historical context is assumed, improving the model's generalized capability.
\end{itemize}

\noindent\textbf{Implementation Details.}
Text conditioning utilizes a fixed global prompt for the description of talking head videos: ``\textit{A person is speaking, and his head moves rhythmically in a small range to follow the sound}''. To optimize memory utilization and ensure numerical stability, we employ a gradient accumulation strategy with steps set to 4, alongside gradient checkpointing~\cite{gradientcheckpointing} for accurate gradient computation.

\subsubsection{Streaming Student Pre-Training}

\label{sec:student_pretraining}

Following the teacher training, we initiate the training of the student model. In this preliminary phase, we adopt a supervised learning approach without introducing knowledge distillation. The primary objective is to adapt the backbone to the streaming generation paradigm.

\noindent\textbf{Training Setup and Optimization Strategy.}
The structural distinction of the student model lies in the integration of the ID-Context Cache mechanism within the self-attention layers, designed to persist Key-Value (KV) pairs of the reference identity across temporal segments. To maintain consistency with the teacher's configuration, the input temporal dimension is fixed at $T_{\text{lat}}=13$ latent frames, corresponding to $T_{\text{raw}}=97$ pixel-space frames. 

We simulate the streaming process by partitioning the latent sequence $\mathbf{Z}(0)$ into $K=4$ sequential chunks. The initial segment comprises 4 latent frames (including one reference frame, $n=4$), while subsequent segments contain 3 frames each, the same as the teacher input.

\noindent\textbf{Sequential Optimization with State Caching.}
Unlike the parallel processing of the teacher, the student model processes chunks sequentially to emulate real-time inference. We apply a noise injection strategy identical to the teacher model. Crucially, the generative latent state of the preceding chunk $\mathbf{z}_{k-1}$ is preserved via the ID-Context Cache to condition the generation of the current chunk $\mathbf{z}_k$. This recursive dependency can be formulated as:
\begin{align}
    \mathbf{h}_k &= \mathcal{S}_{\theta}(\mathbf{c}_k, \mathcal{C}_{\text{cache}}^{(k-1)})\\
    \mathcal{C}_{\text{cache}}^{(k)} &= \text{Update}(\mathcal{C}_{\text{cache}}^{(k-1)}, \mathbf{h}_k)
\end{align}
where $\mathcal{S}_{\theta}$ represents the student network and $\mathcal{C}_{\text{cache}}$ denotes the stored context features. ``Update" denotes the rolling update strategy elaborated in Sec.~\ref{sec:backbone_details}. This mechanism ensures the effectiveness of the streaming generation process.

\noindent\textbf{Implementation Details.}
The optimization employs a learning rate of $\eta = 1 \times 10^{-5}$. Consistent with the teacher training, the pre-trained SpeechAE module undergoes joint fine-tuning with the generative backbone to ensure optimal audio-visual alignment. Furthermore, to guarantee generalization, we strictly adhere to the same dropout regularization strategies and text conditioning prompts utilized in the teacher training phase elaborated in Sec.~\ref{sec:teacher_pretraining}.

\noindent\textbf{Outcome and Limitations.}
Through this streaming pre-training strategy, the student model acquires fundamental streaming generation capabilities. However, empirical observations suggest that while the model learns the basic motion patterns, it exhibits limitations in maintaining high-fidelity identity preservation and long-term temporal consistency, necessitating the subsequent Asynchronous Streaming Distillation (ASD) phase.

\subsubsection{ASD for Streaming Teacher-student Distillation}
\label{sec:asd_training}

Building upon the pre-trained non-streaming teacher and the streaming-adapted student, we implement the Asynchronous Streaming Distillation (ASD) phase. The primary theoretical objective is to distill the global temporal dependency information captured by the teacher into the student model, thereby mitigating the inconsistency and error accumulation issue inherent to the restricted attention window and streaming generation pipeline of the student model.

\noindent\textbf{Noise Alignment and Input Representation.}
To facilitate effective knowledge transfer, we adopt the same segmentation strategy and temporal dimensions ($T_{\text{lat}}=13$) as in the preceding stages. Our primary objective is to maximize the mutual information between the denoised latent vector distributions obtained by the student model and those from the teacher model. This allows the distillation of the teacher’s global temporal dependencies into the student model, which suffers from limited temporal consistency due to constrained attention windows, as detailed in Sec.~\ref{sec:ASD}. Notably, in practice, we reformulate the mutual information constraint on the denoised latent distributions into a mutual information constraint between the predicted velocity flows output directly by the teacher and student models to improve training efficiency. This requires strictly consistent noise sampling between the teacher and student models, which is proved in Sec.~\ref{sec:proveasdcon}.

Accordingly, we share the random seed to sample a global Gaussian noise tensor $\boldsymbol{\epsilon} \sim \mathcal{N}(0, \mathbf{I})$ and apply CANS to generate the noisy input ${\mathit{\boldsymbol{Z}}}(t)$. This ensures that while the signal-to-noise ratios (SNR) vary across asynchronous chunks, the noise structure remains strictly consistent between the teacher and student. The teacher model $\mathcal{T}$ processes the holistic, non-streaming noisy latent sequence $\widetilde{\mathit{\boldsymbol{Z}}}(t) \in \mathbb{R}^{T \times C \times H \times W}$. While the student model $\mathcal{S}$ operates in a streaming manner, receiving the segmented noisy latent sequence:
\begin{equation}
    \widetilde{\mathit{\boldsymbol{Z}}}(t)=[\mathit{\boldsymbol{z}}_\text{R} \mathbin{\|} \mathit{\boldsymbol{Z}}_1(t)\mathbin{\|}\cdots\mathbin{\|}\mathit{\boldsymbol{Z}}_k(t)]
\end{equation}
where $\mathit{\boldsymbol{z}}_\text{R}$ denotes the clean reference image latent. The student performs an iterative chunk-by-chunk denoising process, leveraging the ID-Context Cache to propagate historical states.


\noindent\textbf{Distillation Objectives.}
Our distillation framework imposes constraints directly on the predicted velocity flow outputs, denoted as ${\mathit{\boldsymbol{v}^{\mathcal{T}}}}$ and ${\mathit{\boldsymbol{v}^{\mathcal{S}}}}$. The composite objective function aggregates a fundamental reconstruction target, a mutual information target, and a temporal consistency target:
\begin{equation}
    \mathcal{L}_{\text{ASD}} = \mathcal{L}_{\mathcal{S}_{\theta}}+\alpha\mathcal{L}_{\text{CON}}+\beta\mathcal{L}_{\text{SMO}}
\end{equation}
Detailed theoretical proofs validating the efficacy of these components are provided in Sec.~\ref{sec:proveasdcon}. Briefly, the reconstruction target ($\mathcal{L}_{\mathcal{S}_{\theta}}$) based on Mean Squared Error (MSE) ensures the student's pixel-level fidelity aligns with the teacher's output. The MI target ($\mathcal{L}_{\text{CON}}$) is formulated using the contrastive InfoNCE loss, which maximizes the mutual information between the distributions of the output flows of the teacher and student models. In our implementation, we set $\tau=0.1$ in $\mathcal{L}_{\text{CON}}$. The temporal consistency target ($\mathcal{L}_{\text{SMO}}$) is to address the jitter caused by windowed attention. We minimize the error in the second-order difference of the optical flow dynamics ($\Delta^2$), enforcing smoothness in the generated motion.

In practical engineering implementation, the magnitudes of these loss components differ. To balance their gradients, we empirically set the weighting coefficients to $\alpha=\beta = 0.2$.

\subsubsection{Implementation Details}
During the ASD training phase, the parameters of the teacher model are frozen, allowing it to serve solely as a stable provider of global generative priors. The student model is fully learnable and optimized with a learning rate of $\eta = 1 \times 10^{-5}$. Crucially, we implement Synchronized Dropout Strategies: any dropout mask (Identity, Audio, or Motion Prior) applied to the teacher is identically applied to the student during the forward pass. This synchronization is vital for ensuring valid contrastive learning. As demonstrated in Sec.~\ref{sec:ablation}, the ASD paradigm successfully transfers global temporal coherence to the student, significantly enhancing generation quality, identity preservation, and temporal consistency in streaming scenarios.

\section{Additional Qualitative Results}
In this section, we present a comprehensive and systematic evaluation of the generation quality and stability of our model. Designed as a supplement to the findings in Sec.~\ref{sec:results} of the main paper, this analysis provides deeper insight into the model's robustness and generalization. The evaluation is organized as follows: First, Sec.\ref{sec:suppasd} conducts a fine-grained ablation study specifically targeting our proposed ASD strategy to validate its contribution to generation quality. Next, Sec.\ref{sec:suppactor} rigorously assesses the cross-actor performance of REST, testing the model's generalization capabilities across diverse reference images, diverse age and gender, and diverse driven speech signals. This section serves as a vital complement to the experimental section (Sec.\ref{sec:results}) of the main text. It provides additional empirical evidence that corroborates the rationale and effectiveness of the innovative components in REST. Through these extended experiments, we further demonstrate the model's strong generalization capabilities, thereby substantiating its considerable potential for practical application. For comprehensive qualitative results, please refer to the supplementary video in the Supplementary Materials.

\subsection{Qualitative Ablation Study on ASD}
\label{sec:suppasd}
\begin{figure}[t]
    \centering
    \includegraphics[width=1.0\linewidth]{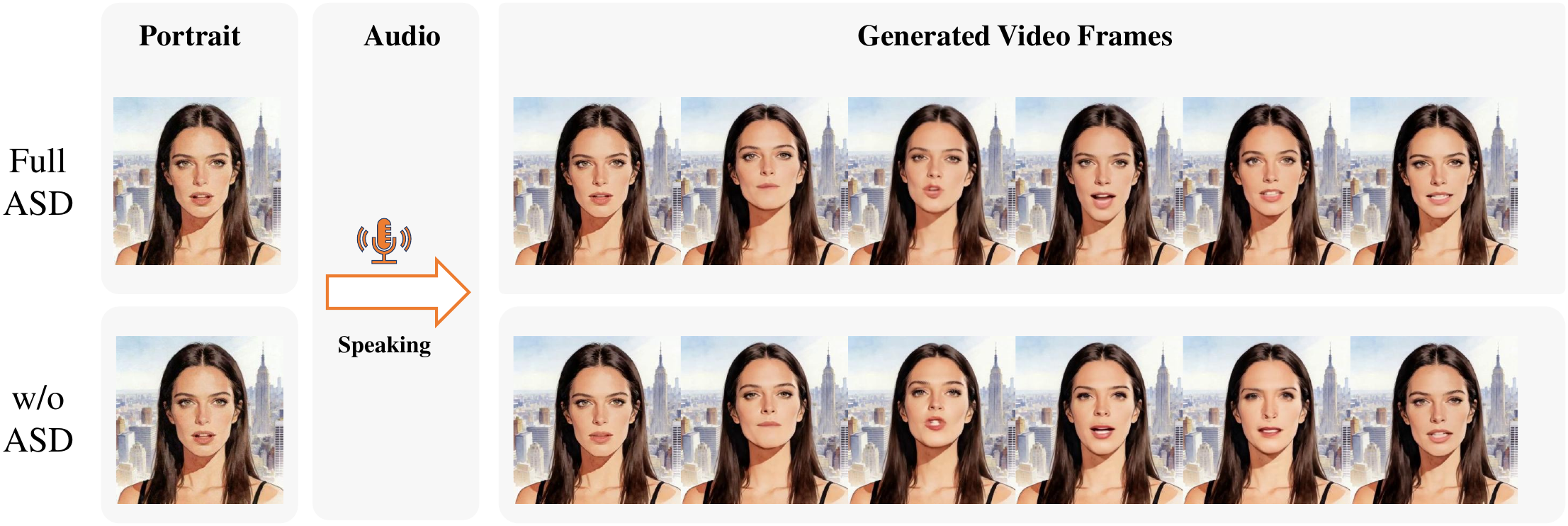}
    \caption{Qualitative Ablation Study on the proposed ASD strategy.}
    \label{fig:suppasd}
\end{figure}

While Sec.\ref{sec:ablation} of the main paper quantitatively demonstrated the efficacy of our Asynchronous Streaming Distillation (ASD) strategy in maintaining ID and temporal consistency, this section presents a complementary qualitative analysis. We aim to provide deeper visual evidence of the superiority of ASD in sustaining generation quality and generalization performance.

To isolate the specific contribution of ASD, we utilize a fixed reference image of a challenging colored pencil portrait style and a consistent female speech audio clip across two distinct experimental configurations:

\begin{itemize}
    \item \textbf{Full ASD:} The proposed method, where the student model undergoes pre-training followed by fine-tuning via ASD.
    \item \textbf{w/o ASD:} A baseline configuration where the model is trained without the ASD strategy, utilizing no knowledge distillation during either pre-training or fine-tuning.
\end{itemize}

To ensure a rigorous and fair comparison, all hyperparameters including training iterations and learning rates are kept identical across both settings. We visualize the generation results by extracting keyframes at synchronized timestamps, as illustrated in Fig.\ref{fig:suppasd}. As observed in the ablation results, the configuration without ASD (\textbf{w/o ASD}) exhibits gradual quality degradation as the streaming inference progresses. This deterioration manifests as increasing blurriness and structural deformation in facial regions, particularly the eyes and lips. This phenomenon serves as a concrete manifestation of the error accumulation inherent in autoregressive streaming models, while our ID-Sink strategy provides partial mitigation, it remains insufficient to fully resolve this issue in isolation. In sharp contrast, configuration without ASD (\textbf{Full ASD}) maintains superior generation quality throughout the streaming inference process, showing no perceptible error accumulation or temporal degradation. These qualitative findings confirm the critical role of ASD in mitigating the error accumulation issue inherent in autoregressive streaming models and preserving long-term ID and temporal consistency, thereby highlighting the significant innovative value of the proposed approach. Please view the supplementary video for the full ablation video.

\subsection{Qualitative Study on Cross-Actor Generalization}
\label{sec:suppactor}

\begin{figure}[t]
    \centering
    \includegraphics[width=1.0\linewidth]{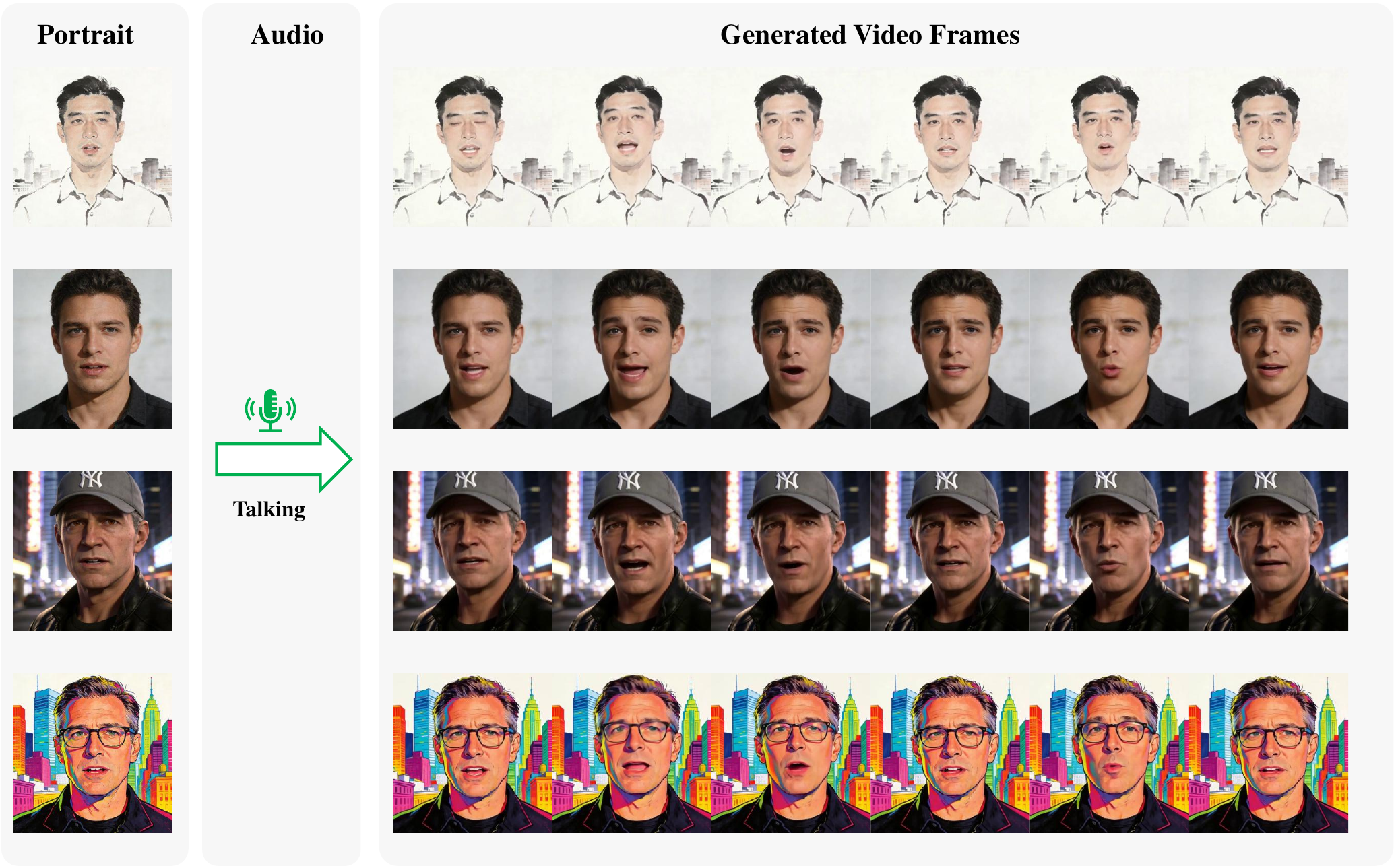}
    \caption{Qualitative Study on Cross-Actor Generalization of male subjects.}
    \label{fig:suppactor1}
\end{figure}

\begin{figure}[t]
    \centering
    \includegraphics[width=1.0\linewidth]{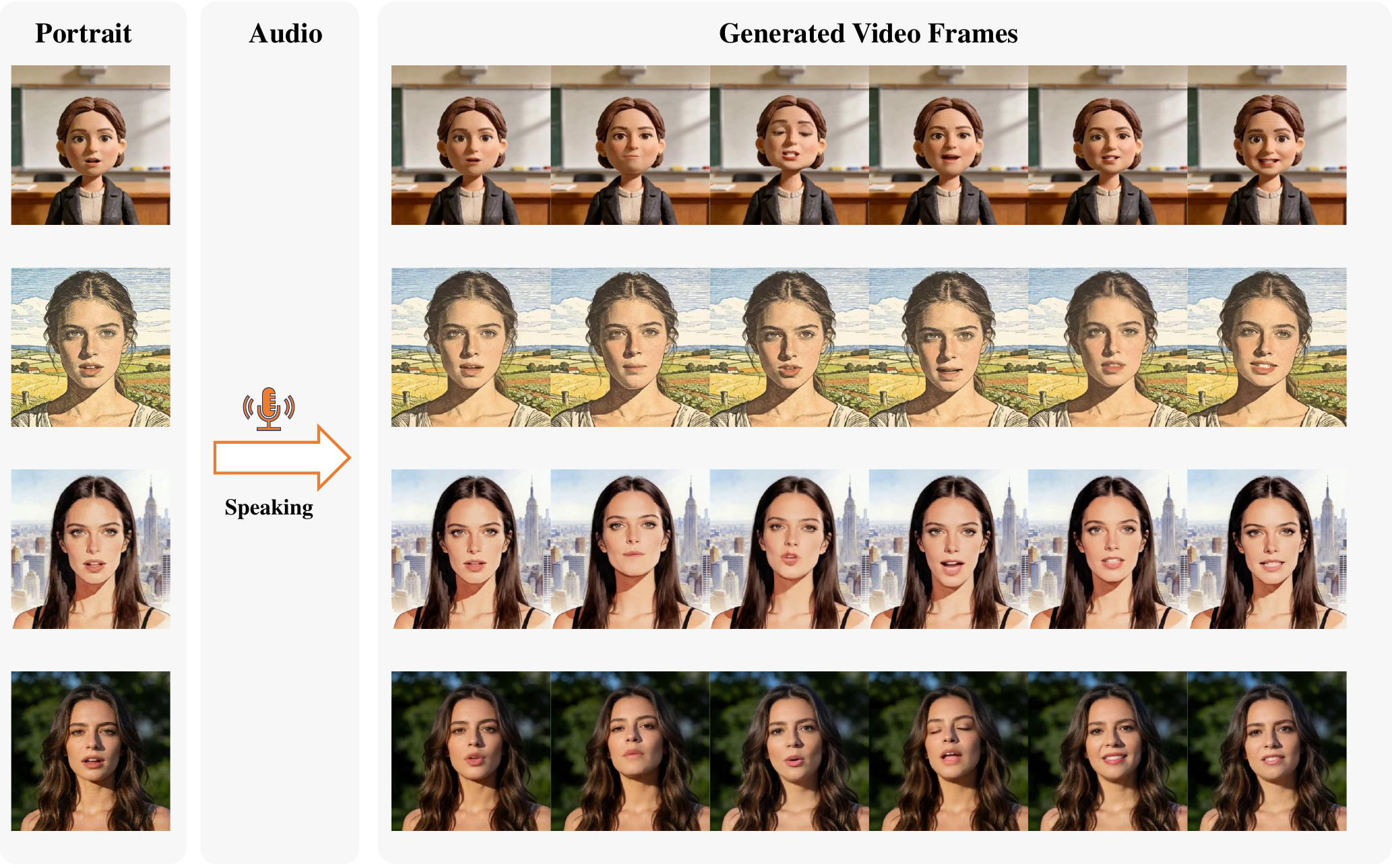}
    \caption{Qualitative Study on Cross-Actor Generalization of female subjects.}
    \label{fig:suppactor2}
\end{figure}

Generalization capability across diverse reference images, audio sources, and demographics (gender and age) constitutes a critical benchmark for evaluating talking head generation models, serving as a primary indicator of their practical application potential. In this subsection, we conduct a systematic evaluation of the proposed REST framework regarding its robustness to varying stylistic and acoustic conditions. To rigorously assess universality, we categorize our evaluation into two scenarios:
\begin{itemize}
    \item \textbf{Male Subject Scenarios:} We utilize audio from classic cinematic footage as the driving signal to test robustness against background noise and varied prosody. The reference images include four representative styles: \textit{ink wash painting}, \textit{close-up photography}, \textit{CG animation}, and \textit{fluorescent painting}.
    \item \textbf{Female Subject Scenarios:} We employ a speech recording characterized by natural reverberation to test acoustic generalization. The reference images encompass \textit{cartoon figurine}, \textit{woodblock painting}, \textit{colored pencil sketch (hand-drawing)}, and \textit{real-person photography} styles.
\end{itemize}

This comprehensive setup allows us to meticulously measure the model's performance across distinct artistic modalities and complex acoustic environments. We perform a visual analysis by extracting keyframes at synchronized timestamps across all generated sequences. The results are illustrated in Fig.~\ref{fig:suppactor1} and Fig.~\ref{fig:suppactor2}, with full video provided in the supplementary material.

The visualization results demonstrate that our proposed model exhibits strong generalization capabilities. First, when provided with driver reference images of different genders and artistic styles, our proposed model consistently generates high-quality outputs that maintain faithful ID preservation across all styles, highlighting the effectiveness of our proposed methods in ensuring ID consistency. Second, the model achieves robust driving performance across diverse input portrait styles. Notably, at identical key timestamps, the synthesized lip shapes for different styles are highly consistent. This observation further confirms the model's excellent generalization performance when applied to portraits of varying styles. Furthermore, the REST model maintains accurate audio-visual synchronization not only with clean, cinematic speech but also with real-world speech containing reverberation. This demonstrates the model's robustness to different types of driven speech inputs. In summary, our proposed REST framework demonstrates exceptional ID consistency and driving robustness when conditioned on reference images of varying styles, genders, and ages, as well as on diverse speech types. These results underscore REST's superior generalization ability and its strong potential for practical, real-world applications.

\section{Limitations and Future Work}
\label{sec:futurework}

Despite the significant advancements REST achieves in real-time, expressive audio-driven talking head generation, the framework encounters specific technical challenges that delineate avenues for future investigation.

First, while REST produces realistic head motions, it occasionally exhibits motion blur artifacts, particularly during rapid or large-amplitude head movements. This issue stems primarily from limitations in the training data, especially the presence of motion-blurred samples corresponding to rapid or extensive head rotations. To mitigate this, future iterations could employ two key strategies: (1) refining the training distribution by rigorously excluding samples with excessive optical flow magnitudes; and (2) integrating automated motion blur detection algorithms to pre-filter low-fidelity videos before training.

Second, another limitation is the occasional lack of clarity in fine-grained facial structures, specifically within the dental region. This can be attributed to two factors: insufficient lip and teeth clarity in the original dataset, and the aggressive spatial compression strategy of our temporal VAE, which may discard fine-grained oral details. To address this, we propose two potential solutions: (1) applying super-resolution techniques to the training data to enhance the ground-truth fidelity of dental textures; and (2) implementing a semantically-guided loss function that utilizes oral segmentation masks to assign higher weights to the mouth and teeth regions, thereby explicitly penalizing reconstruction errors in these critical areas.

Notwithstanding these limitations, REST constitutes a creative and effective framework that successfully balances real‑time streaming capability with high expressive quality in talking head generation. Its design offers a meaningful trade‑off between inference speed and visual fidelity, marking an important milestone toward practical and responsive digital human interaction systems. We believe REST provides a foundational framework that will significantly accelerate the deployment and research of real-time talking head video generation.

\section{Ethical Consideration}
\label{sec:ethicalconsideration}

The REST framework demonstrates the capability to synthesize photorealistic talking head videos with high perceptual fidelity. While this technological advancement unlocks transformative potential across domains, ranging from empathetic human-computer interaction and remote education to virtual caregiving companionship, we unequivocally acknowledge the dual-use nature of such high-fidelity generation. The ability to produce indistinguishable synthetic media presents inherent risks of malicious exploitation. These risks include the fabrication of disinformation involving public figures, the generation of non-consensual explicit content, and the creation of deceptive media for fraudulent purposes. Such misuse fundamentally contravenes our research objective, which is to harness generative AI for societal betterment.

To mitigate these risks, we have integrated a multi-layered ethical safety protocol spanning the entire development of REST:

\noindent\textbf{Data Curation and Sanitization.} 
Ethical AI development begins with the training data. During the data preparation phase, we enforced rigorous filtering protocols to sanitize the training corpus. We systematically excluded any samples containing violence, sexual themes, or otherwise inappropriate semantic content. This ensures that the model does not learn to generate harmful artifacts from its source distribution.

\noindent\textbf{Restricted Deployment and Oversight.} 
We have implemented strict access controls governing the deployment of REST. Currently, the model is restricted to a research-only environment under the direct supervision of our internal risk assessment team. A stringent manual review process is applied to all inputs, including both image and audio, to preemptively block the generation of malicious content. For any prospective public release, we commit to establishing a rigorous auditing framework to guarantee that the generated outputs remain benign and legally compliant.

\noindent\textbf{Advocacy for Digital Forensics.} 
Beyond immediate safeguards, we strongly advocate for the parallel advancement of deepfake detection technologies. The development of robust forensic methods to identify synthetic media is a critical community-wide imperative. We view the improvement of generation quality and the advancement of forgery detection as coupled research goals. Both are essential for mitigating the societal risks associated with generative media and preventing the potential misuse of REST.

Through an analysis of the potential societal implications of our proposed REST framework and the corresponding mitigation strategies we have adopted, we aim to collaborate actively with the community to ensure its use within controlled and ethical boundaries. By implementing the above measures to prevent technological misuse, we strive to advance the development of generative artificial intelligence while safeguarding against its potential abuse in areas such as deepfake generation.

\section{Summary}
This supplementary document provides a comprehensive and detailed exposition of the theoretical foundations and practical implementations of our proposed REST model, serving as an integral supplementation to the main manuscript. 

Sec.~\ref{sec:dataset} supplements the content of Sec.~\ref{sec:results} in the main paper. It introduces the training and evaluation datasets used in our experiments, elaborates on their characteristics, justifies the rationale behind their selection, and provides a complete description of our data preprocessing pipeline.

Sec.~\ref{sec:architecture} offers a detailed technical breakdown of the REST model architecture. This section provides an in-depth analysis of our novel ID-Context Cache module and a thorough description of the core Streaming DiT backbone, thereby serving as a direct supplement to Sec.~\ref{sec:idcontext} and Sec.~\ref{sec:ASD} of the main text.

Sec.~\ref{sec:traindetails} extensively details the training and inference procedures of the REST model. It includes a theoretical analysis of the Asynchronous Streaming Distillation (ASD) paradigm and provides mathematical proofs for the validity of the designed loss functions. Furthermore, this section documents essential implementation details, including weight initialization strategies, hardware specifications, and the exact hyperparameter configurations used in our experiments, thus complementing the discussions in Sec.~\ref{sec:ASD} and Sec.~\ref{sec:results} of the main manuscript.

Sec.~\ref{sec:futurework} critically assesses the current limitations of the REST framework. By analyzing these constraints, we identify open challenges and propose potential trajectories for future research in audio-driven talking head generation.

Sec.~\ref{sec:ethicalconsideration} concludes the document by addressing the ethical considerations and social impacts associated with our technology.

In summary, this supplementary document enriches the theoretical and practical framework of the proposed REST model, providing essential guidance for its real-world application and ensuring the reproducibility of the experimental results.

\end{document}